\documentclass{article}

\usepackage{microtype}
\usepackage{graphicx}
\usepackage{subfigure}
\usepackage{booktabs} 

\usepackage{hyperref}

\usepackage[accepted]{icml2020}

\icmltitlerunning{AdaSGD: Bridging the gap between SGD and Adam}

\usepackage[utf8]{inputenc} 
\usepackage[T1]{fontenc}    
\usepackage{url}            
\usepackage{booktabs}       
\usepackage{amsfonts}       
\usepackage{nicefrac}       
\usepackage{microtype}      
\usepackage[dvipsnames]{xcolor}
\usepackage{graphicx}
\usepackage{amsmath,amsfonts,amssymb,amsthm}
\usepackage{xcolor,colortbl}
\usepackage{mathtools}
\usepackage{multirow}
\usepackage{rotating}
\usepackage{pifont}

\newcommand{\xmark}{\ding{55}}

\DeclarePairedDelimiter{\ceil}{\lceil}{\rceil}
\newtheorem{theorem}{Theorem}[section]
\newtheorem{corollary}{Corollary}[theorem]

\begin{document}

\twocolumn[
\icmltitle{AdaSGD: Bridging the gap between SGD and Adam}



\icmlsetsymbol{equal}{*}

\begin{icmlauthorlist}
\icmlauthor{Jiaxuan Wang}{umich}
\icmlauthor{Jenna Wiens}{umich}
\end{icmlauthorlist}

\icmlaffiliation{umich}{Department of Computer Science and Engineering, University of Michigan}


\icmlkeywords{Machine Learning, ICML}

\vskip 0.3in
]



\printAffiliationsAndNotice{}  

\title{AdaSGD: Bridging the gap between SGD and Adam}

\begin{abstract}
In the context of stochastic gradient descent (SGD) and adaptive moment estimation (Adam), researchers have recently proposed optimization techniques that transition from Adam to SGD with the goal of improving both convergence and generalization performance. However, precisely how each approach trades off early progress and generalization is not well understood; thus, it is unclear when or even if, one should transition from one approach to the other. In this work, by first studying the convex setting, we identify potential contributors to observed differences in performance between SGD and Adam. In particular, we provide theoretical insights for when and why Adam outperforms SGD and vice versa. We address the performance gap by adapting a single global learning rate for SGD, which we refer to as AdaSGD. We justify this proposed approach with empirical analyses in non-convex settings. On several datasets that span three different domains, we demonstrate how AdaSGD combines the benefits of both SGD and Adam, eliminating the need for approaches that transition from Adam to SGD. 
\end{abstract}

\section{Introduction}

Many machine learning tasks require the choice of an optimization method. In general, this choice affects not only how long it takes to reach a reasonable solution, but also the generalizability of that solution. In recent years, many adaptive gradient methods, such as AdaGrad \cite{duchi2011adaptive}, RMSProp \cite{tijmen2012}, and Adam \cite{kingma2015adam}, have been proposed. Such methods, which adapt the learning rate for each parameter, have become the benchmark in many applications. In particular, Adam is widely used, since in practice its default learning rate works well across many problems \cite{karpathy2019,harutyunyan2017multitask,xu2015show,oh2017value}. However, there remain settings in which state-of-the-art performance is achieved by applying SGD with momentum \cite{cubuk2018autoaugment,gastaldi2017shake,huang2018gpipe}. Given these observed differences, researchers have started to investigate transition rules to switch from Adam to SGD \citep{luo2019adaptive,keskar2017improving}. However, there are still gaps in our understanding of \textit{when} and \textit{why} Adam outperforms SGD and \textit{vice versa}. Consequently, it remains unclear as to whether or not transitional approaches are even required.

In this paper, we present a rigorous empirical comparison of SGD with momentum (referred to as SGD throughout for simplicity) and Adam. We aim to increase our understanding of the strengths and weaknesses of each approach, so that we can make informed choices. 
Through theoretical analyses in the convex setting and a series of experiments on both synthetic and real non-convex data, we demonstrate that by adapting a single global learning rate within SGD, the gap between SGD and Adam can be reduced. The contributions are summarized below.

\begin{itemize}
    \item To build intuition, we show how and why Adam and SGD can fail in the least squares regression setting and then extend this intuition to SGD with logistic and exponential loss.
    \item Based on the intuition, we propose a simple modification of SGD, AdaSGD, that results in many of the good properties enjoyed by SGD and/or Adam both theoretically (in the convex setting) and empirically (using deep networks of various architectures across domains). The modification isolates the contribution of adapting the learning rate on a per parameter basis versus globally.
    \item Compared to previously proposed transitional methods from Adam to SGD, AdaSGD has fewer hyperparameters and we show that it is more robust to hyperparameter selection. 
\end{itemize}

 The rest of the paper is organized as follows. Following a review of related work (Section \ref{related_work}), we introduce our proposed approach in Section \ref{AdaSGD}. Then, we identify desirable properties of Adam and SGD on the synthetic data (Section \ref{quadratic_loss}) and verify those properties on real data (Section \ref{real_datasets}).
Overall, this work demonstrates that a simple modification to SGD can greatly reduce the gap between SGD and Adam, without introducing new hyperparameters to tune, obviating the need for transitional approaches. 

\section{Preliminaries and Background} \label{related_work}

Before presenting our empirical and theoretical findings, we first introduce the problem setup and notation used throughout. Then, we give a brief overview of related work, providing additional context. 

\textbf{Problem setup.} We consider a scenario in which the goal is to optimize some objective function $\min_{\boldsymbol{\theta}} \mathbb{E}_{(\boldsymbol{x},y)\sim \mathcal{D}}  L(\boldsymbol{x}, y; \boldsymbol{\theta})$ for some distribution $\mathcal{D}$ and loss function $L$. Given $n$ labeled training examples, where $\{(\boldsymbol{x}^{(i)}, y^{(i)}): i \in [1, \cdots, n] \}$ are sampled $i.i.d.$ from $\mathcal{D}$, we aim to minimize empirical risk $\frac{1}{n}\sum_{i=1}^n L(\boldsymbol{x}^{(i)}, y^{(i)};\boldsymbol{\theta})$.  Throughout, we use bold font to denote vectors.

In this work, we first build intuition on the differences between Adam and SGD, by restricting $L$ to be a strongly convex quadratic function. Then, we consider settings where $L$ corresponds to the non-convex loss associated with the output of a deep neural network,  verifying intuition established from solving a convex optimization problem.

\textbf{Optimization Algorithms.} We focus on comparing SGD and Adam, two of the most commonly used optimization algorithms in machine learning (and specifically deep learning \cite{bottou2018optimization}). SGD updates its model parameters according to the gradient of the loss with respect to a randomly sampled example ($\boldsymbol{x}_t, y_t$) from the training set: $\boldsymbol{\theta}_{t+1} = \boldsymbol{\theta}_{t} - \eta \nabla_{\boldsymbol{\theta}} L(\boldsymbol{x}_t, y_t \ ; \ \boldsymbol{\theta}_{t}) \ $, where $\boldsymbol{\theta}_{t} \in \mathbb{R}^d$ is the parameter vector at iteration $t$, $\eta \in \mathbb{R}^{+}$ is the learning rate, and $d$ is the dimensionality of the feature vector. For brevity, we shorthand $L(\boldsymbol{x}_t, y_t \ ; \ \boldsymbol{\theta}_{t})$ with $L(\boldsymbol{\theta}_{t})$. Following common practice, we add momentum to SGD. 
$$\text{\emph{SGD with momentum: }} \boldsymbol{\theta}_{t+1} = \boldsymbol{\theta}_{t} - \eta m(\nabla_{\boldsymbol{\theta}} L( \boldsymbol{\theta}_{t})) \ $$

where $m(\boldsymbol{g}_t) = \beta_1 m(\boldsymbol{g}_{t-1}) + \boldsymbol{g}_t$ with $m(\boldsymbol{g}_0) = \boldsymbol{0}$. Throughout, we refer to SGD with momentum as SGD. 

In contrast, Adam adapts the learning rate for each parameters separately according to the following update rule:
$$\text{\emph{Adam: }} \begin{aligned}\boldsymbol{\theta}_{t+1} = \boldsymbol{\theta}_{t} - \eta \frac{\mathbb{E}_{\beta_1}(\nabla_{\boldsymbol{\theta}} L(\boldsymbol{\theta}_{t}))}{\sqrt{\mathbb{E}_{\beta_2}([\nabla_{\boldsymbol{\theta}} L(\boldsymbol{\theta}_{t})]^2)} + \epsilon} \frac{\sqrt{1-(\beta_2)^t}}{1-(\beta_1)^t}\end{aligned} \ $$

with element-wise division. $\mathbb{E}_{\beta}(\boldsymbol{w}_t) = \beta \mathbb{E}_{\beta}(\boldsymbol{w}_{t-1}) + (1-\beta) \boldsymbol{w}_t$ is the $0$ initialized exponential average ($\mathbb{E}_{\beta}(\boldsymbol{w}_0)=\boldsymbol{0}$) with a non-negative hyperparameter scalar $\beta$. $\beta_1$, $\beta_2$, $\epsilon$ are also hyperparameters, typically set to some default
. The last term corrects bias due to $0$ initialization. This adaptation scheme can be interpreted as approximating second order methods with a diagonal Hessian. 

\textbf{Convergence Properties.} 
For many years, researchers have studied the convergence properties of SGD \cite{bubeck2015convex,bottou2018optimization,ruder2016overview}. Notably, in a convex setting, SGD has a regret bound that depends on $d$, whereas AMSGrad \cite{reddi2019convergence}, a variant of Adam, does not. Instead, AMSGrad's bound depends on $\sum_{i=1}^d \Vert \boldsymbol{g}_{1:T, i} \Vert_2$. This suggests that AMSGrad may converge faster when the gradient is sparse. We will derive a similar bound for our proposed method to show convergence. 
Extending beyond the convex setting, recently, researchers have started to analyze the convergence properties of adaptive methods in non-convex settings \cite{chen2018convergence, zhou2018convergence, li2018convergence}, and empirically have  demonstrated that adaptive methods have a faster initial progress during training. However, given a standard training budget, SGD is often able to catch up towards the end \cite{wilson2017marginal, luo2019adaptive, chen2018convergence}. Here, we focus on progress made near the end of the training, using a training budget explored in previous work 
(training curves are included in \textbf{Appendix B}). 



\textbf{Generalization Performance.} Empirically, researchers have observed that in some cases SGD produces solutions that generalize better compared to Adam \citep{luo2019adaptive,keskar2017improving,wilson2017marginal,loshchilov2017fixing}. One explanation is that SGD is algorithmically stable (\textit{i.e.,} the solution is robust to perturbations of the training data) \cite{hardt2015train}. \textbf{We will show that Adam is not}. Moreover, in the least squares regression setting, \textit{each step} in gradient descent (GD), starting from the origin, corresponds to a point in the parameter space that approximates a solution with some amount of $L2$ regularization \cite{goodfellow2016deep}. \textbf{We empirically show that SGD follows the same path but Adam does not}. Furthermore, if the linear model is over-determined and initialized in the span of the data, SGD converges to the minimum $L2$ norm solution, unlike Adam \cite{zhang2016understanding, reddi2019convergence}.


\textbf{Closing the gap.} Previously, researchers have shown that the performance gap between SGD and Adam can be closed if one tunes some default hyperparameters of Adam. For example, with large $\epsilon$, Adam becomes SGD and thus it is not surprising that tuning $\epsilon$ for each problem would bridge the gap \cite{choi2019empirical}. Similarly, instead of taking the square root of the exponential average for the square of the gradient, a smaller exponent (say $1/8$ instead of $1/2$) brings Adam closer to SGD (note: Adam is equivalent to SGD when the exponent is 0) and is shown to perform well on computer vision tasks \cite{chen2018closing}. Instead of tuning more hyperparameters, which can be expensive, \textbf{we aim to understand why the gap exists and can we bridge the gap with minimal tuning}. Our explanation offers a different point of view on why tuning $\epsilon$ and the exponent helps.



\textbf{Transitional Methods.} Another line of work that aims to close the gap proposes methods that transition from Adam to SGD \cite{luo2019adaptive,keskar2017improving}, with the goal of leveraging the faster initial progress of Adam, before switching to SGD for a more generalizable solution. AdaBound \cite{luo2019adaptive} and Swats \cite{keskar2017improving} are two `transitional' approaches with different criteria for \textit{when} to switch. AdaBound uses a `soft' transition scheme. It defines an upper bound function, $\eta_u (t) = \eta_{\text{sgd}} \cdot (1 + 1/(\gamma \cdot t))$, and a lower bound function, $\eta_l(t) = \eta_{\text{sgd}} \cdot (1 - 1/(\gamma \cdot t + 1))$, on the current iteration number $t$ to clip the adaptive learning rate within the bounds with parameters $\gamma$, $\eta_{\text{sgd}} \in \mathbb{R}^{+}$. When $t=0$, the bound is loose $(0, \infty)$, so AdaBound initially behaves like Adam. As $t$ approaches $\infty$, $\eta=\eta_{\text{sgd}}$, so the algorithm converges to SGD. Unlike AdaBound, Swats determines $\eta_{\text{sgd}}$ and the switching point automatically. It starts with Adam and switches to SGD when the magnitude of changes in Adam's update projected onto the direction of the gradient stabilizes. It then uses this stabilized learning rate as $\eta_{\text{sgd}}$ and transitions to SGD. Though both approaches have been shown to work in certain settings, \textbf{we will show in this work that Adam is \textit{not} always faster than SGD}. Thus, starting with Adam and then switching to SGD may not be beneficial. Moreover, as highlighted above, AdaBound requires specifying additional hyperparameters: the learning rate for Adam (before transition) $\eta_{\text{adam}}$, the learning rate for SGD (after transition) $\eta_{\text{sgd}}$, and the switching point $\gamma$. On real datasets, we compare the performance of Swats and Adabound to SGD, Adam, and our proposed approach. Moreover, we compare their robustness to hyperparameter selection.  


\section{Proposed approach -- AdaSGD} \label{AdaSGD}

As Adam and SGD have been found to be favorable in different settings, we provide a middle ground, AdaSGD, that combines features from both. Our proposed method, AdaSGD, combines SGD's property of implicit regularization and Adam's ability to adapt step sizes to the loss surface. The main idea involves adapting a \textit{global} learning rate for SGD, increasing its robustness across problem settings.  Previous work provides a rich theoretical motivation for step size adaptation, especially for adapting a global learning rate \cite{vaswani2019painless, li2018convergence}. For example, Li and Orabona analyzed the converegence of a global step size adaptation of SGD based on AdaGrad in non-convex settings. In contrast, we base our adaption on Adam due to its popularity and show empirically that it works better than Li's approach across datasets (\textbf{Appendix A}).  
Proofs for the properties of our proposed approach are included in \textbf{Appendix E}.

\textbf{Update rules:} AdaSGD uses the following rules to adapt the learning rate and update model parameters:
$$\text{\emph{AdaSGD: }}\boldsymbol{\theta}_{t+1} = \boldsymbol{\theta}_{t} - \eta_t m(\nabla_{\boldsymbol{\theta}} L(\boldsymbol{\theta}_{t})) $$ 

where $m$ is the momentum function, $\eta_t = \eta \frac{\sqrt{1-(\beta_2)^t}}{\sqrt{v_t / d}}$ is a scalar, $v_t = \mathbb{E}_{\beta_2}(\Vert \nabla_{\boldsymbol{\theta}} L(\boldsymbol{\theta}_{t})\Vert_2^2)$, $v_0=0$, and $\eta > 0$ is a hyperparameter. Note that AdaSGD is different from Adam in that it adapts only the global learning rate $\eta$ instead of adapting each parameter's learning rate. That is for one dimensional problems, AdaSGD and Adam are almost equivalent (`almost' because AdaSGD uses momentum while Adam uses exponential average of the gradient; we don't expect this to make a difference in practice). We adapt the learning rate based on Adam as a mechanism to explore the differences between SGD and Adam; We do not claim the adaptation scheme to be optimal.  We note that AdaSGD is a special case of AdaShift \cite{zhou2018adashift} (without the decorrelation operation, with a global learning rate, and uses the $L2$ function to aggregate second moment estimation spatially, instead of the max used in AdaShift). These simplifications allow us to pinpoint the potential reason that Adam performs poorly in some settings, namely fitting to directions associated with small eigenvalues. While we’d expect layer-wise adaptation (as used in AdaShift) to better cope with different statistics across layers, normalization techniques could solve the problem for deep networks even though AdaSGD only uses a global learning rate.

For the same reason that Adam fails to converge, AdaSGD does not converge with convex loss. Thus, to better understand the properties of AdaSGD, we use the same trick introduced in \citealp{reddi2019convergence}, \textit{i.e.}, we analyze a close variant AdaSGDMax instead. In AdaSGDMax, we have $\eta_t = \eta / {\sqrt{\hat{v}_t / d}}$ where $\hat{v}_t = \max\{ \hat{v}_{t-1}, v_t / (1-(\beta_2)^t) \}$ and $\hat{v}_{0} = 0$. In the case of constrained optimization, \textit{i.e.}, $\forall t$, $\boldsymbol{\theta}_t \in \mathcal{F} $ where $\mathcal{F}$ is some constrained set, we project $\boldsymbol{\theta}_t$ to $\Pi_{\mathcal{F}}(\boldsymbol{\theta}_t) := \min _{\boldsymbol{y} \in \mathcal{F}} \Vert \boldsymbol{\theta}_t - \boldsymbol{y}\Vert_2$ to satisfy the constraint. 

\begin{theorem}[\textbf{AdaSGDMax is robust to the choice of learning rate}] \label{theorem:robust_to_lr}
With strongly convex deterministic quadratic loss and bounded gradient, AdaSGDMax with $\beta_1 = 0$ converges for all $\eta>0$ unless $\lim_{t \to \infty}\eta_t = \frac{2} {\lambda_{\max}}$.
\end{theorem}

AdaSGDMax greatly expands SGD's convergent range. The exceptional case can be easily avoided by decaying the learning rate once the loss stops decreasing. Note that SGD with any learning rate decay that has a positive final learning rate in the convergent range will converge regardless of the initial learning rate. However, a learning rate that is too small could lead to slow convergence. A nice property of AdaSGDMax, as shown in the proof, is that it will not decrease the learning rate once $\eta_t$ is in SGD's convergent range, preventing it from suffering slow convergence due to a small learning rate. A byproduct of the proof is that AdaSGDMax converges linearly (like SGD). It spends little time in the non-convergent region of SGD.

\begin{theorem}[\textbf{AdaSGDMax has sublinear regret}]
Consider a constrained online optimization problem with a convex constraint set $\mathcal{F}$,  convex loss function $f_t$ with $\Vert \nabla f_t(\boldsymbol{\theta}) \Vert_\infty \leq G_\infty$ for all $t \in [T]$ and $\boldsymbol{\theta} \in \mathcal{F}$, we aim to minimize the regret $R_T := \sum_{i=1}^T f_t(\boldsymbol{\theta}_t) - \min_{\boldsymbol{\theta}^* \in \mathcal{F}} \sum_{i=1}^T f_t(\boldsymbol{\theta}^*)$. Assuming $\mathcal{F}$ has bounded diameter $D_\infty$ (\textit{i.e.,} $D_\infty = \max_{\boldsymbol{x}, \boldsymbol{y} \in \mathcal{F}} \Vert \boldsymbol{x} - \boldsymbol{y} \Vert_{\infty})$ and $\eta_t = \eta / \sqrt{t \hat{v}_t / d}$ 
AdaSGDMax has regret 
$$R_T \leq \frac{D^2_\infty \sqrt{d \hat{v}_T T}}{2 \eta} + \frac{d^{3/2} G_\infty^2 \eta (2 \sqrt{T}-1)}{2 \sqrt{\hat{v}_1}}$$
\end{theorem}

\begin{corollary}
Setting $\eta_t = \eta D_\infty / (G_\infty \sqrt{t \hat{v}_t })$, we have
$$R_T \leq \frac{d D_\infty G_\infty \sqrt{\hat{v}_T T}}{2 \eta} + \frac{d D_\infty G_\infty \eta (2 \sqrt{T}-1)}{2 \sqrt{\hat{v}_1}}$$
\end{corollary}

Here, we study the same online optimization problem as in \citealp{reddi2019convergence} and \citealp{luo2019adaptive}. Sublinear convergence in this setting implies convergence in the empirical risk minimization setting \cite{reddi2019convergence}. Note that our bound is comparable to SGD's regret bound of $\frac{d D_\infty G_\infty \sqrt{T}}{2\eta} + \frac{d D_\infty G_\infty \eta(2\sqrt{T}-1)}{2}$ \cite{hazanintroduction}. When $\hat{v}_T$ is small and $\hat{v}_1$ is large, AdaSGDMax can make faster progress compared to SGD. When the gradient is sparse, however, AMSgrad (a variant of Adam that is guaranteed to converge with convex loss) can converge faster. However, we will show that such scenarios rarely occur in the least squares regression setting. Moreover, in a non-convex setting using real data, given a standard number of training epochs, Adam does not lead to a lower training loss compared to AdaSGD.  


\textbf{AdaSGD v.s. AdaSGDMax:}
The only difference between AdaSGD and AdaSGDMax is that in the former the learning rate can increase. Empirically, this leads to good solutions faster, since the algorithm can accelerate in flat regions and slow down when the gradient changes quickly. The next theorem shows that even with the ability to increase the learning rate, AdaSGD has bounded error for deterministic strongly convex quadratic problems.

\begin{theorem}[\textbf{AdaSGD reaches a solution close to the optimal solution}] \label{theorem:close_to_sln_text}
The distance from AdaSGD's solution $\boldsymbol{\theta}$ to the optimal solution $\boldsymbol{\theta}^*$ for deterministic strongly convex quadratic problems with $\eta > 0$ and bounded gradient is bounded by $\Vert \boldsymbol{\theta} - \boldsymbol{\theta}^* \Vert_2 \leq \frac{\sqrt{d} \eta \mathcal{K}} { 2 (1-\beta_2)}$, where $\mathcal{K}=\lambda_{\max} / \lambda_{\min}$ is the condition number.
\end{theorem}

A consequence of Theorem \ref{theorem:close_to_sln_text} is that decaying the learning rate will lead AdaSGD to a more accurate solution. 

\section{Lessons from convex quadratic loss} \label{quadratic_loss}

\begin{table}[t]
    \centering
    \begin{tabular}{c|cc}
         & Implicit regularization & Robustness to $\eta$ \\
         \midrule
         SGD &  \checkmark & \xmark \\
         Adam & \xmark & \checkmark \\
         AdaSGD & \checkmark & \checkmark 
    \end{tabular}
    \caption{A comparison among SGD, Adam, and AdaSGD.}
    \label{tab:properties}
\end{table}

\begin{figure*}
\centering
\subfigure[]{\label{fig:angle_contour}
\includegraphics[width=0.17\linewidth]{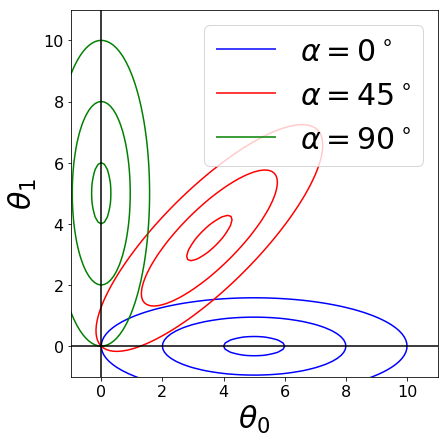}}
\subfigure[]{\label{fig:angle}
\includegraphics[width=0.25\linewidth]{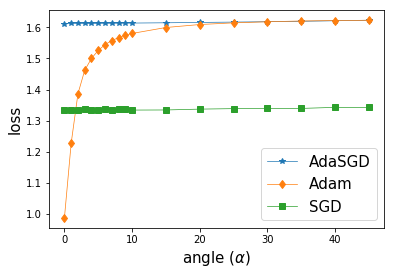}}
\subfigure[]{\label{fig:lambda_adaSGD} 
  \includegraphics[width=0.24\linewidth]{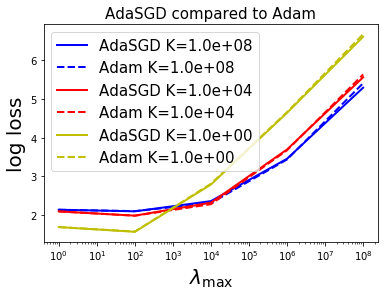}}
\subfigure[]{\label{fig:angle_dimension}
  \includegraphics[width=0.27\linewidth]{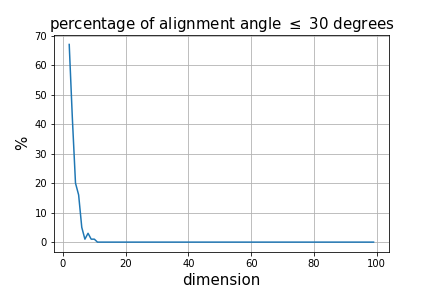}}
\caption{Adam has an advantage when the Hessian of the problem is nearly diagonal (which is a rare event). \textbf{(a)} Angle alignment illustration with three differently colored contour plots, and \textbf{(b)} comparison of different optimization approaches applied to problems with varying angle alignment on 2D data with $\mathcal{K}=10^4$, $\lambda_{\min}=1$, $\eta_{sgd}=1/\lambda_{\max}$, $\eta_{AdaSGD}=0.0005$, and $\eta_{Adam} = 0.005$.
\textbf{(c)} AdaSGD with a single learning rate ($\eta=0.01$) has similar loss compared to Adam with a single learning rate ($\eta=0.1$). \textbf{(d)} A near diagonal Hessian becomes increasingly rare as the number of dimensions increases.}
\label{fig:angles}
\end{figure*}

Before examining the performance on real datasets and deep networks, we first explore properties of Adam, SGD, and AdaSGD in a simpler setting using synthetic data. Following previous work \cite{wilson2017marginal,zhang2016understanding}, we consider a setting in which one aims to minimize a convex quadratic loss function. While deep networks generate more complex loss surfaces, a convex quadratic loss is a good local approximation. In addition, as we will show, using a convex quadratic loss, one can reproduce much of the phenomena observed in training deep networks using SGD or Adam. Further theoretical insights for SGD that extend beyond quadratic loss are also discussed in this section. 

We begin by formally introducing the problem setup specific to quadratic loss and our data generation process. For settings in which Adam has favorable properties compared to SGD, we show that AdaSGD has similar properties. In parallel, for settings in which SGD has favorable properties compared to Adam, we show that AdaSGD maintains those properties. \textbf{Table \ref{tab:properties}} provides an overview of the properties.

\textbf{Least squares regression.} 
Here, we minimize the squared loss over the training set $L(\boldsymbol{\theta}) = \frac{1}{2} \Vert X \boldsymbol{\theta} - \boldsymbol{y} \Vert_2^2$, where $X \in \mathbb{R}^{n \times d}$ is the design matrix. Note that $X^\intercal X = Q^\intercal \Lambda Q$ for an orthonormal matrix $Q$ and a diagonal matrix $\Lambda$ (spectral theorem). Furthermore, all entries of $\Lambda$ are non-negative. We denote $\lambda_{\max}$ and $\lambda_{\min}$ as the largest and smallest eigenvalues of $X^\intercal X$ respectively (that is the largest and smallest diagonal entries of $\Lambda$). If $X^\intercal X$ is invertible (diagonal entries of $\Lambda$ are all positive), then the optimal solution is $\boldsymbol{\theta}^* = (X^\intercal X)^{-1} X^\intercal \boldsymbol{y}$. The condition number, a proxy for the difficulty of the task, is denoted as $\mathcal{K} := \lambda_{\max} / \lambda_{\min}$. In the following sections, we will explore how Adam and SGD perform on this task as we vary the condition number. 

\textbf{Evaluation.} Since both Adam and SGD converge slowly with a large condition number \cite{ali2017}, instead of measuring time to convergence, we compare their performance by measuring loss after a fixed number of parameter updates ($3,000$), well after the loss stops changing for problems with small condition numbers. The results when sweeping the number of parameter updates is in \textbf{Appendix B} and are consistent with the conclusions drawn in this section. Furthermore, since different tasks have different minimum attainable losses, to account for task difficulty, we report the regret in loss, \textit{i.e.} $L(\boldsymbol{\theta}) - L(\boldsymbol{\theta}^*)$, and refer to it as `loss' throughout this section.

\textbf{Data generation.} To explore situations in which SGD outperforms Adam and vice versa, we generate synthetic data for which we know the true underlying solution, $\boldsymbol{\theta}^*$. As described above, we focus on least squares.  To generate a least squares problem, we first specify $X$ and $\boldsymbol{y}$, where $X = V \Sigma Q$, with $V$ and $Q$ being orthonormal matrices, and $\Sigma \in \mathbb{R}^{n \times d}$ being a diagonal matrix with $\Sigma^{\intercal} \Sigma =\Lambda$. Since we are interested in how performance varies with different $\mathcal{K}$, we vary $\lambda_{\max}$  and $\lambda_{\min}$. To generate a diverse set of least squares problems, given $\lambda_{\max}$  and $\lambda_{\min}$, we fill the diagonal entries of $\Lambda$ by placing $\lambda_{\max}$ at the first entry,  $\lambda_{\min}$ at the last entry, and selecting the remaining values from a uniformly spaced log scale. We sample $Q$ and $V$ uniformly in the orthogonal group $O(n)$ and $\boldsymbol{y} \sim \mathcal{N}(0, 30)^d$, inducing a relationship between $X$ and $\boldsymbol{y}$.  We initialize parameters $\boldsymbol{\theta}_0 \sim \mathcal{N}(0,1)^d$. Following convention, we set $\beta_1=0.9$ and $\beta_2=0.999$. We set $n=300$ and $d=100$. For both SGD and Adam, at each iteration, we uniformly sample a single data point to compute an estimate of the gradient. 

\subsection{When Adam has an edge over SGD} \label{Adam_better}


Compared to SGD, Adam often leads to a faster initial decrease in training loss \cite{wilson2017marginal} and has a default learning rate that works well across problem settings. 

\underline{\textbf{4.1.1 Advantage Adam:}} Adam achieves lower loss with a fixed number of iterations compared to SGD when the Hessian associated with the problem is diagonal.

\underline{\textbf{Justification:}}
Since Adam adapts each parameter's learning rate separately, we expect quick convergence for problems with a diagonal Hessian, \textit{i.e.}, when the axes of the ellipsoidal contour associated with the quadratic problem are aligned with the standard basis. \textbf{Figure \ref{fig:angle_contour}} illustrates the angle of alignment $\alpha$. As expected, when we vary $\alpha$ from $0$ to $45$ degrees ($45$ to $90$ degrees is a mirror image) in \textbf{Figure \ref{fig:angle}}, Adam performs the best when $\alpha$ is close to $0$ degree.

\underline{\textbf{Solution - AdaSGD:}} Admittedly, AdaSGD does not close the gap between SGD and Adam on problems with a diagonal Hessian, but a near diagonal Hessian becomes increasingly rare as $d$ increases (\textbf{Figure \ref{fig:angle_dimension}}). Here, we define the alignment angle of a unit vector $\boldsymbol{v}$ as $cos^{-1}(\Vert \boldsymbol{v} \Vert_{\infty})$. Then, for a uniformly random unit vector (a row of $Q$), as we increase $d$, the probability of having a large coordinate (small alignment angle) decreases exponentially. Even when one considers all eigenvectors (rows of a uniform randomly generated orthonormal matrix $Q$), as $d$ increases, nearly no eigenvectors are axis-aligned. 

\underline{\textbf{4.1.2 Advantage Adam:}} Adam's learning rate is robust to varying $\lambda_{\max}$, whereas SGD's is not.

\begin{figure*}
\centering
\subfigure[]{\label{fig:lambda_adam}
\includegraphics[width=0.3\linewidth]{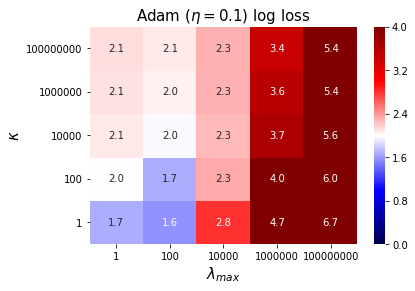}}
\subfigure[]{\label{fig:lambda_SGD}
\includegraphics[width=0.3\linewidth]{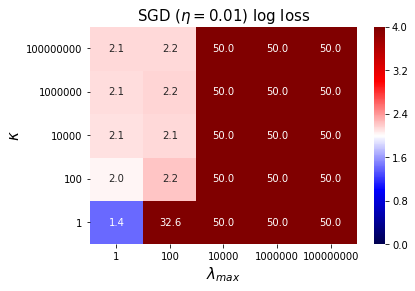}}
\subfigure[]{\label{fig:lambda_optSGD}
\includegraphics[width=0.3\linewidth]{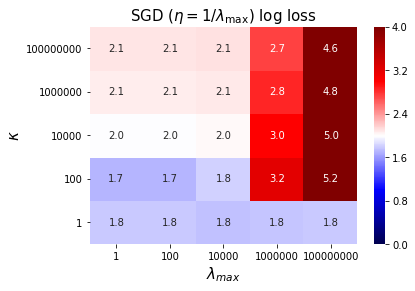}}
\caption{We randomly generated least squares problems with $\lambda_{\max}$ and $\mathcal{K}$ in the range $[1, 10^8]$ and $d=100$. For each setting of $\lambda_{\max}$ and $\mathcal{K}$, we averaged training loss over $30$ different randomly generated datasets with $n=300$. We used $\eta=0.1$ for Adam, and $\eta=0.01$ for SGD. These learning rates were chosen to minimize average loss across tasks. Here, we show log loss plot of \textbf{(a)} Adam ($\eta=0.1$), \textbf{(b)} SGD ($\eta=0.01$), and \textbf{(c)} SGD ($\eta = \frac{1}{\lambda_{\max}}$). $50$ denotes diverged solutions. SGD is not robust as we vary $\lambda_{\max}$, but with an appropriate learning rate, it can be just as fast as Adam. 
}
\label{fig:lambda}
\end{figure*}

\underline{\textbf{Justification:}}
A big advantage of using Adam is that a single learning rate \cite{karpathy2019} achieves similar losses in many settings given conventional training budget, whereas SGD does not (\textbf{Figure \ref{fig:lambda}}). This phenomena is not solely due to SGD diverging with a large learning rate (which can be solved by learning rate decay) as even in the convergent regions, SGD with a single learning rate is far from optimal (comparing \textbf{Figure \ref{fig:lambda}b} and \textbf{\ref{fig:lambda}c}). 




\underline{\textbf{Solution - AdaSGD:}} SGD is not inherently slower than Adam. By making SGD's learning rate problem dependent, in this case $\eta= 1/\lambda_{\max}$, SGD achieves a loss similar to the loss achieved by Adam after a fixed number of iterations (\textbf{Figure \ref{fig:lambda_optSGD}}). In contrast, AdaSGD can find such a problem dependent learning rate automatically, which is desirable in scenarios with unknown $\lambda_{\max}$ or more complex settings. As shown in \textbf{Figure \ref{fig:lambda_adaSGD}}, AdaSGD achieves a training loss comparable to Adam. This result suggests that at least with convex quadratic loss, there is little advantage in adapting each dimension's learning rate separately. This increased robustness to learning rate is supported by Theorem \ref{theorem:robust_to_lr}.

\subsection{When SGD has an edge over Adam} \label{SGD_better}


Here, we show how both SGD and AdaSGD benefit from implicit regularization, whereas Adam does not. In particular, SGD has a close connection with the minimum norm solution in the least squares regression setting. Furthermore, for many commonly used loss functions (\textit{i.e.}, quadratic loss, logistic loss, and exponential loss), as we will show, SGD's update rule has an intuitive connection to principal component analysis (PCA). This parallel also connects SGD for training a deep neural network with nonlinear PCA, suggesting that SGD implicitly performs dimensionality reduction. 

\begin{figure}
\centering
\subfigure[]{\label{fig:sub1}
\includegraphics[width=0.45\linewidth]{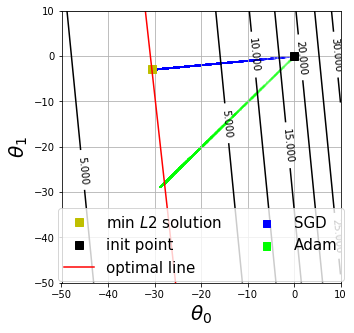}}
\subfigure[]{\label{fig:sub2}
\includegraphics[width=0.51\linewidth]{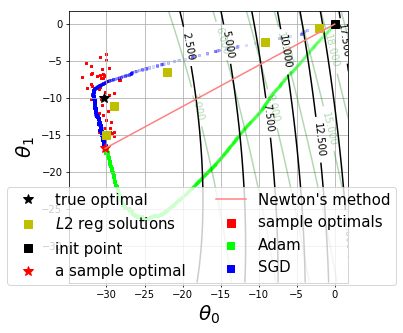}}
\caption{An illustration of why SGD may lead to better generalization performance compared to Adam. \textbf{(a)} With multiple global minimums, SGD initialized in the row space of $X$ chooses the minimum norm solution. \textbf{(b)} SGD, initialized at $0$, follows the path of $L2$ regularization solutions. A complete description of the plot is included in Section \ref{SGD_better}.} 
\label{fig:fig1}
\end{figure}


\underline{\textbf{4.2.1 Advantage SGD:}} SGD initialized in the row space of $X$ converges to the minimum $L2$ norm solution when $X^\intercal X$ is not invertible \cite{zhang2016understanding,wilson2017marginal}.

\underline{\textbf{Justification}}:
We illustrate this phenomenon through \textbf{Figure \ref{fig:sub1}}, where $\lambda_{\min}=0$ (because $X^\intercal X$ is not invertible). The loss contours are parallel due to the extra degree of freedom caused by the dependent columns of $X$. SGD takes a direct path to the solution. In contrast, Adam magnifies the gradient along the y-axis because of its smaller size, and thus, drifts along the $0$ eigenvalue direction (the red line). 

\underline{\textbf{Solution - AdaSGD:}} AdaSGD also does not update in the directions associated with the $0$ eigenvalue, thus converging to the minimum $L2$ norm solution. 

\underline{\textbf{4.2.2 Advantage SGD:}} SGD's optimization path corresponds to different $L2$ regularization values, whereas Adam's optimization path does not. 

\underline{\textbf{Justification:}} \label{opt_path_explanation}
Not only does SGD have an implicitly regularized solution, the solution path of SGD is also implicitly regularized. For deterministic quadratic loss with $0$ initialized model parameters, SGD with early stopping is equivalent to $L2$ regularization \cite{goodfellow2016deep}. In \textbf{Figure \ref{fig:sub2}}, we show that this relation approximately holds for a stochastic setting as well (each yellow block corresponds to a solution with a different regularization strength). Adam, on the other hand, takes an opposite route when approaching the solution. It tends to increase the learning rate along directions associated with small eigenvalues, and thus misses the true solution. Here, we generate $300$ data points randomly according to our data generation procedure described earlier, with $\lambda_{\min}=1$ and $\lambda_{\max}=10$. Then, we randomly sample $10$ data points as training data. The small red dots correspond to the optimal solutions for $50$ different random samples. The black star is the optimal point for all $300$ data points (black lines are the loss contour for the true optimal point, whereas green lines correspond to the loss contour for a random sample). Along the direction associated with the smallest eigenvalue, the variance of the solution is the greatest.  
In this case, although SGD is slower in arriving at the training solution (Newton's method is the fastest as it converges in one step), if stopped early, it leads to better generalization performance compared to Adam and Newton's method. 

\underline{\textbf{Solution - AdaSGD:}} AdaSGDMax follows the same optimization path as SGD, once in SGD's convergent range. 
With $\eta \ll 1$, in the deterministic strongly convex quadratic setting, we have $Q (\boldsymbol{\theta}_{t+1} - \boldsymbol{\theta}^*) = (I - \eta_t \Lambda) Q (\boldsymbol{\theta}_t - \boldsymbol{\theta}^*)$. That is, the error goes to $0$ the fastest along the direction associated with the largest eigenvalue, just like SGD. 


\textbf{Why is fitting small eigenvalue directions bad for generalization?} Empirical evidence on MNIST and CIFAR-10 is given in \textbf{Appendix C}. Here, we give two explanations, one related to $L2$ regularization and the other related to PCA. First, observe that with regularization strength $\alpha > 0$, $Q \boldsymbol{\theta}_{L2} = (\Lambda + \alpha I)^{-1} \Lambda Q \boldsymbol{\theta}^*$, where $\boldsymbol{\theta}_{L2}$ is the regularized solution and $\boldsymbol{\theta}^*$ is the unregularized solution. That is, the solution along the smallest eigenvalue direction is suppressed compared to an unregularized solution. Given the success of $L2$ regularization, we prefer to follow directions associated with large eigenvalues. 

Second, for the least squares setting, if the data are zero centered, the eigenvectors with large eigenvalues are the principal components of $X^\intercal X$. In other words, SGD prioritizes optimizing high variance directions. This intuition carries over if we treat all but the last layer (assumed to be linear) of a deep neural network as feature transformations. The principal components identified at the last layer define a data manifold when viewed in the input space, effectively corresponding to nonlinear PCA. Fortunately, this phenomena is not tied specifically to squared loss. Consider a binary classification problem (\textit{i.e.}, $Y \in \{-1,1\}$) and define $v_i=y^{(i)} \mathbf{\theta}^\intercal \mathbf{x}^{(i)}$. For logistic loss, the Hessian becomes $X^\intercal \Gamma_{log} X$, with the $i^\text{th}$ entry of the diagonal matrix $\Gamma_{log}$ being $\frac{e^{v_i}}{(1+e^{v_i})^2}$. Observe that due to the squared term in the denominator, $\Gamma_{log}$ places more weight on samples with larger loss (\textit{i.e.}, $v_i < 0$). Similarly, for exponential loss, the Hessian is $X^\intercal \Gamma_{exp} X$, with the $i^\text{th}$ entry of the diagonal matrix $\Gamma_{exp}$ being $e^{-v_i}$. Again, more weight is placed on misclassified points. 

The message is clear: with logistic and exponential loss, SGD prioritizes optimizing along the components with the greatest variance with respect to `misclassified' points. One should note that this interpretation requires the data to be zero centered and that the Hessian does not differ much along the path between the current point and the optimal point. The first requirement can be justified with data normalization, while the second is an approximation that Newton's method also assumes \cite{bradie2006friendly}.

\section{Closing the gap on real datasets} \label{real_datasets}

\begin{table*}[t]
  \caption{A comparison of optimization algorithms. In the first section of the table, we display results that correspond to a full grid search for learning rate $\eta$ and $L2$ regularization strength $\alpha$. In the second section of the table, we show how AdaSGD is more robust to a single learning rate compared to SGD. In parentheses we include 95\% bootstrapped confidence intervals. For each dataset, we bolded the best result and underlined the worst result in each section.}
  \centering
    \begin{tabular}{ccccc}
    Hyperparameters & Methods & MIMIC (AUC) & CIFAR-10 (accuracy) & WikiText-2 (perplexity) \\
    \midrule\\
    \multirow{5}{*}{Tuned}
    &SGD & \textbf{0.852} (0.839, 0.870) & 94.20 (93.74, 94.62) & 123.52 (121.89, 125.26) \\
    &Adam & 0.847 (0.825, 0.863) &  93.43 (92.96, 93.88) & \textbf{108.37} (106.69, 109.89) \\
    &AdaSGD & 0.849 (0.826, 0.868) & 94.00 (93.52, 94.46) &  109.96 (108.38, 111.56) \\
    &Swats & \underline{0.846} (0.826, 0.862) & \underline{93.29} (92.83, 93.76) & 122.77 (120.81, 124.42) \\
    &AdaBound & 0.850 (0.825, 0.866) & \textbf{94.82} (94.37, 95.25) & \underline{267.43} (263.26, 271.19) \\
    \midrule
    \multirow{3}{*}{Fixed}
    &SGD ($\eta=0.1$) & \underline{0.841} (0.820, 0.859) & \textbf{94.20} (93.76, 94.63) & \underline{171.60} (168.95, 174.10) \\
    &AdaSGD ($\eta=10^{-4}$) & \textbf{0.849} (0.825, 0.869) & 93.87 (93.40, 94.33) & \textbf{109.96} (108.38, 111.56) \\
    &Adam ($\eta=10^{-3}$) & 0.847 (0.827, 0.869) &  \underline{93.43} (92.95, 93.89) & 123.69 (121.99, 125.46) \\
    \midrule
    \end{tabular} \label{table:result_table}
\end{table*}

Based on the insights gained from synthetic data, we explore the benefits of AdaSGD in more realistic settings, using three datasets across domains. We compare the performance of SGD, Adam, AdaSGD, with transitional methods (AdaBound and Swats), and  demonstrate that transitional methods add little benefit compared to AdaSGD. 

\begin{figure*}
\centering
\subfigure[]{\label{fig:cifar_final_lr}
\includegraphics[width=0.38\linewidth]{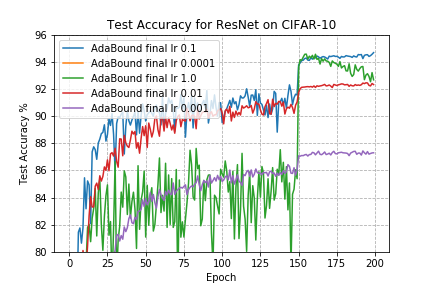}}
\subfigure[]{\label{fig:wiki2_transition}
\includegraphics[width=0.37\linewidth]{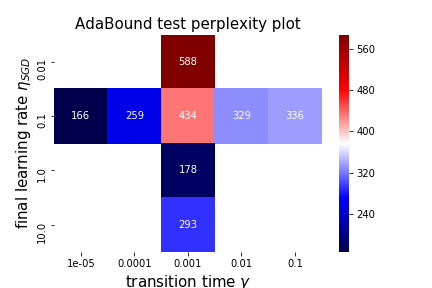}}
\caption{\textbf{(a)} CIFAR-10: AdaBound is sensitive to the final learning rate, $\eta_{\text{sgd}}$. \textbf{(b)} WikiText-2: AdaBound is sensitive to the final learning rate and the transition scheme $\gamma$.}
\label{fig:transition_scheme}
\end{figure*}



\subsection{Experimental setup} \label{exp_setup}

We consider three datasets: MIMIC-3 \cite{johnson2016mimic,harutyunyan2017multitask}, CIFAR-10 \cite{krizhevsky2009learning}, and WikiText-2 \cite{merity2016pointer}, selected because they are common benchmarks within their respective domains, namely healthcare, computer vision, and natural language processing. For MIMIC-3, we convert the pipeline in \cite{harutyunyan2017multitask} from Keras to PyTorch. For CIFAR-10, we use code provided in \cite{luo2019adaptive}. For WikiText-2, we adapt code provided in the PyTorch tutorial on language modeling. 
The architectures used are a 2 layer LSTM \cite{hochreiter1997long}, a ResNet, and a 1 layer LSTM respectively. We initialize $\boldsymbol{\theta}_0$ with PyTorch's default initialization \cite{paszke2017automatic}. We include our anonymized code in the \textbf{Supplementary material}. 

In each experiment, we use the validation set to tune the $L2$ regularization and the learning rate. The search ranges are included in \textbf{Appendix A}. 
We run our experiments on the MIMIC-3 dataset with 5 different random seeds and experiments on the other two datasets with 3 random seeds. The reported results are for a model selected based on validation performance. Note that since CIFAR-10 does not have a validation set, we randomly fix 20\% of the training data as the validation data. Based on previous work and our preliminary results, we run an LSTM on MIMIC-3 for $60$ epochs. We run ResNet on CIFAR-10 for $200$ epochs and decay its learning rate by a factor of $10$ at epoch $150$. We run an LSTM on WikiText-2 for $15$ epochs decaying its learning rate by $10$ at epoch $10$. Decaying the learning rate helps the algorithm stabilize around the solution.

 For MIMIC-3, we report the area under the receiver operating characteristic curve (AUC). For CIFAR-10, we report accuracy. For WikiText-2, we report the perplexity of the learned language model (lower is better).

\subsection{Results \& Discussion}

\textbf{Table \ref{table:result_table}} summarizes results. We group them into two sections. The first section displays results when we performed a full grid search tuning both the learning rate and $L2$ regularization. The second section presents results with a fixed learning rate selected based on validation performance across datasets. We observe the following.

\textbf{AdaSGD closes the gap when Adam performs better than SGD.}  On the WikiText-2 dataset, AdaSGD clearly closes the gap when Adam outperforms SGD. AdaSGD and Adam yield comparable performance on the three datasets, when the learning rate and regularization strength are tuned via grid search (upper half of the table).

\textbf{While SGD is sensitive to the learning rate, AdaSGD and Adam are more robust.} When we use a fixed (\textit{i.e.}, not problem specific) learning rate with decay, SGD's performance varies relative to its performance when the learning rate is tuned in a problem specific manner. Tuning the learning rate for AdaSGD leads to only a minimal improvement in performance compared to using a single learning rate. This implies that Adam's robust learning rate is \textit{not} due to its ability to adjust the learning rate separately for each dimension, since adjusting $\eta$ alone (as is the case in AdaSGD) can achieve a similar effect.



\textbf{Transitional methods do not appear to have a consistent edge over their prototypes.} Comparing AdaBound and Swats to SGD and Adam (all tuned for the learning rate and the regularization strength, and setting remaining hyperparameters to their defaults), it is not clear that transitional methods have any advantage. While Swats has the worst performance on two of the datasets, AdaBound's performance also varies. While AdaBound achieves the best accuracy on CIFAR-10, it also performs the worst on the WikiText-2 dataset. The reason for this lower performance is due in part to its sensitivity to both $\eta_{\text{sgd}}$ and $\gamma$. As shown in \textbf{Figure \ref{fig:cifar_final_lr} and \ref{fig:wiki2_transition}}, AdaBound's default hyperparameters (the center block in \textbf{Figure \ref{fig:wiki2_transition}}) are far from optimal.

In addition to the three datasets reported here, we also ran baselines on the Tiny ImageNet dataset, experimented with a 100-layer densenet on the CIFAR-100 dataset, and applied a variational autoencoder on the MNIST dataset. The results are consistent with our finding above, and are documented in (\textbf{Appendix D}.

\section{Conclusion}

Using a convex quadratic loss to build intuition, we demonstrated both empirically and theoretically how SGD benefits from implicit regularization, while Adam does not. We also showed how Adam is more robust to the choice of learning rate compared to SGD. Based on these observations, we proposed a variant of SGD, AdaSGD. By adapting a global learning rate, AdaSGD combines the benefits of both Adam and SGD with minimal tuning. On synthetic and real datasets, we showed how AdaSGD can close the gap in performance between Adam and SGD. Moreover, this simple yet effective change yielded performance that is on par if not better than recently proposed `transitional' approaches.

We note a few limitations of this study. First, though empirical results are encouraging, we do not claim that AdaSGD is the optimal way to adapt the learning rate. Second, in settings where one can identify important dimensions and align the axis of the parameter space with the eigenvectors of those dimensions, then Adam could still have an edge over AdaSGD (\textit{i.e.}, converge faster). 
 \cite{wilson2017marginal}. Nonetheless, AdaSGD helps bridge the gap between these two commonly used stochastic optimization approaches.

\bibliography{ref.bib}
\bibliographystyle{icml2020}

\newpage
\phantom{blabla}

\newpage
\appendix
\section{Hyperparameter search range and result for \cite{li2018convergence}}

In this work, we tuned hyperparameters $\eta$ and $\lambda$ in the following ranges. 

$\eta_{\text{sgd}} \in \{10,1,0.1,0.01,0.001\}$,
$\eta_{\text{adasgd}} \in \{0.0003, 0.0001, 0.00003, 0.00001, 0.000003\}$, $\eta_{\text{adabound}} \in \{0.01, 0.001, 0.0001, 0.00001, 0.000001\}$, $\eta_{\text{swats}} \in \{0.01, 0.003, 0.001, 0.0003, 0.0001\}$, 
$\eta_{\text{adam}} \in \{0.01, 0.003, 0.001, 0.0003, 0.0001\}$

$\lambda_{\text{sgd}} \in \{10^{-7},10^{-6}, 10^{-5}, 10^{-4}, 10^{-3}\}$, $\lambda_{\text{adasgd}} \in \{10^{-7},10^{-6}, 10^{-5}, 10^{-4}, 10^{-3}\}$, $\lambda_{\text{adabound}} \in \{0.0001, 0.003, 0.001, 0.003, 0.01\}$, $\lambda_{\text{swats}} \in \{10^{-7}, 3 \cdot 10^{-7}, 10^{-6}, 3 \cdot 10^{-6}, 10^{-5} \}$, 
$\lambda_{\text{adam}} \in \{10^{-7}, 3 \cdot 10^{-7}, 10^{-6}, 3 \cdot 10^{-6}, 10^{-5}\}$

None of the results reported used hyperparameters at the boundaries of the search ranges.

\section{Training curves across datasets}

\begin{figure*}
\centering
\subfigure[]{
\includegraphics[width=0.24\linewidth]{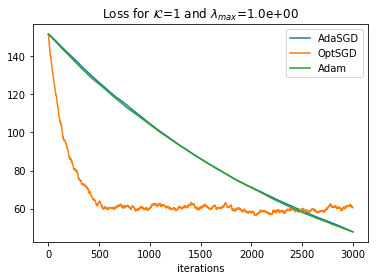}}
\subfigure[]{
\includegraphics[width=0.24\linewidth]{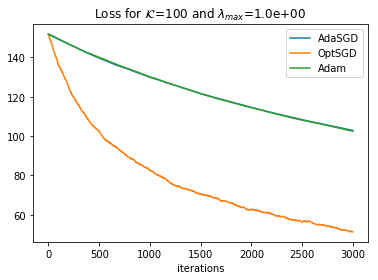}}
\subfigure[]{
\includegraphics[width=0.24\linewidth]{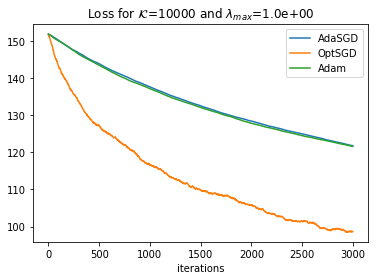}}
\subfigure[]{
\includegraphics[width=0.24\linewidth]{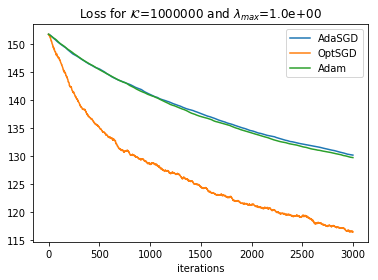}}
\subfigure[]{
\includegraphics[width=0.24\linewidth]{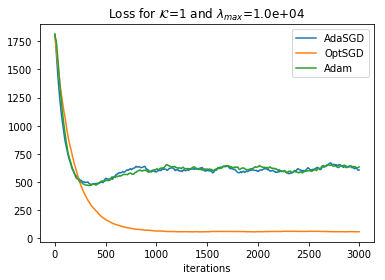}}
\subfigure[]{
\includegraphics[width=0.24\linewidth]{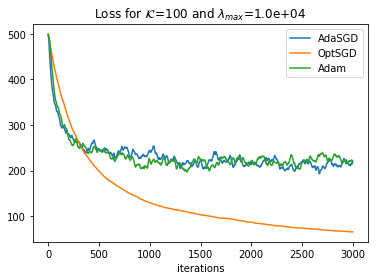}}
\subfigure[]{
\includegraphics[width=0.24\linewidth]{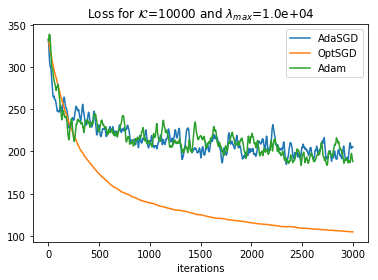}}
\subfigure[]{
\includegraphics[width=0.24\linewidth]{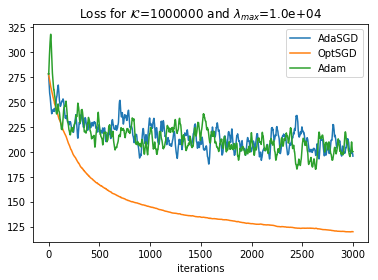}}
\caption{Training curves averaged over $30$ runs for experiments in \textbf{Section 4}. Note that across settings, AdaSGD and Adam are comparable on every epoch (our choice of $3,000$ updates used in \textbf{Section 4} is chosen arbitrarily). Further notice that when SGD is set to the optimal learning rate, it converges faster than both Adam and AdaSGD.}
\label{fig:synthetic_curve}
\end{figure*}

\begin{figure*}
\centering
\subfigure[]{
\includegraphics[width=0.32\linewidth]{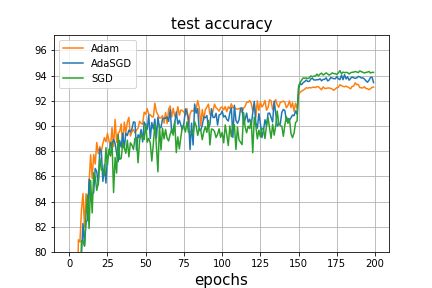}}
\subfigure[]{
\includegraphics[width=0.32\linewidth]{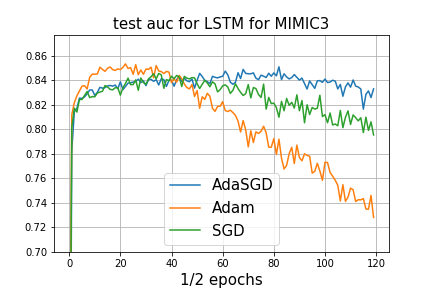}}
\subfigure[]{
\includegraphics[width=0.32\linewidth]{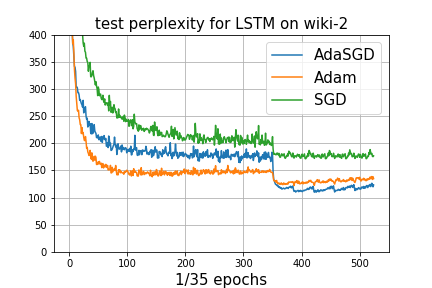}}
\caption{Test accuracy across training epochs for the fixed learning rate experiment in \textbf{Section 5} with \textbf{(a)} CIFAR-10, \textbf{(b)} MIMIC3 and \textbf{(c)} wikitext-2 datasets. Note that despite the initial faster progress of Adam, AdaSGD would eventually catch up the performance gap within conventional training epochs.}
\label{fig:real_curve}
\end{figure*}

In this section, we show the training curves for synthetic datasets in \textbf{Figure \ref{fig:synthetic_curve}} and training curves for the real datasets in \textbf{Figure \ref{fig:real_curve}}. On the synthetic datasets, unless the Hessian is diagonal, there's no difference in the convergence speed between AdaSGD and Adam. On the real datasets, given the number of epochs for training reported in previous work \citealp{harutyunyan2017multitask} for MIMIC3, \citealp{luo2019adaptive} for CIFAR-10, and PyTorch's language modeling tutorial \footnote{\url{https://github.com/yunjey/pytorch-tutorial/blob/master/tutorials/02-intermediate/language_model/main.py}}, AdaSGD performs on par with SGD and Adam at the end of training.

\section{Fitting in directions associated with small eigenvalues could lead to poor generalization}


Stable algorithms lead to better generalization \cite{bousquet2002stability}. Here we show that for a convex quadratic loss, directions associated with small eigenvalues are not stable, so fitting them will likely lead to overfitting. In particular, we show that swapping out a single point in the training data leads to a large change in the solution (as measured by Euclidean distance and change in loss) along the directions associated with small eigenvalues.

We use the notation $X^{-i}$ to denote the replacement of the $i^{th}$ row of the design matrix with a different data point drawn $i.i.d$ from $\mathcal{D}$. We are interested in the change in the solution as $X$ changes into $X^{-i}$ (denote the solutions as $\boldsymbol{\theta}$ and $\boldsymbol{\theta}^{-i}$ respectively). We measure the change in the solution in the basis of the eigenvectors as $Q (\boldsymbol{\theta} - \boldsymbol{\theta}^{-i})$. This choice of basis is natural, because we can link changes in solution to eigenvalues.

\begin{figure*}
\centering
\subfigure[]{\label{fig:change_sub1}
\includegraphics[width=0.24\linewidth]{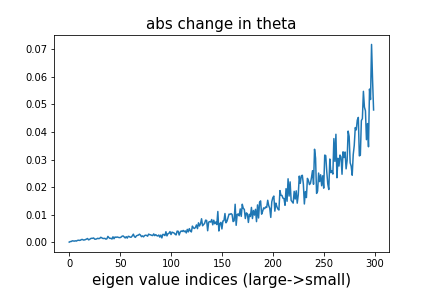}}
\subfigure[]{\label{fig:change_sub2}
\includegraphics[width=0.24\linewidth]{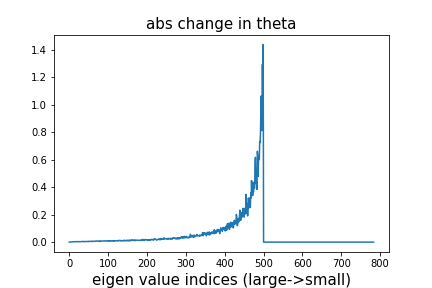}}
\subfigure[]{\label{fig:change_sub3}
\includegraphics[width=0.24\linewidth]{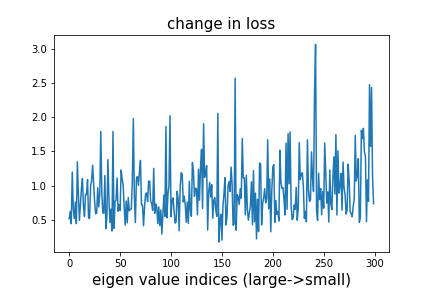}}
\subfigure[]{\label{fig:change_sub4}
\includegraphics[width=0.24\linewidth]{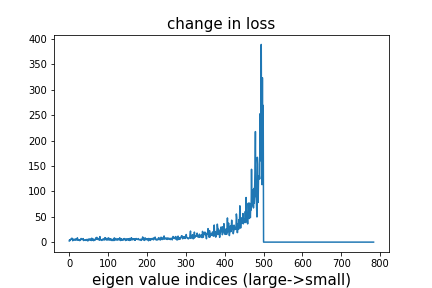}}
\caption{Absolute changes in parameters along each eigenvector's direction for \textbf{(a)} CIFAR-10 and \textbf{(b)} Fashion MNIST. Note that for Fashion MNIST, eigenvalue indices of $500$ and above have eigenvalues of $0$. A similar trend is observed for loss on \textbf{(c)} CIFAR-10 and \textbf{(d)} Fashion MNIST.}
\label{fig:change_fig1}
\end{figure*}

To test our hypothesis on real data, we randomly sampled $500$ data points from the CIFAR-10 and Fashion MNIST datasets \cite{xiao2017fashion}. The swapping procedure is done by randomly choosing a data sample from the sampled points and replacing it with a new sample from the corresponding original dataset. We then solve for $\boldsymbol{\theta}$ and $\boldsymbol{\theta}^{-i}$ for the sampled and swapped dataset, treating them as least squares problems (in the degenerative case, \textit{i.e.} $\lambda_{\min}=0$, we solve for the minimum $L2$ norm solution). We then report $|Q (\boldsymbol{\theta} - \boldsymbol{\theta}^{-i})|$ averaged for $10$ random swappings. Note that the result is a vector. We present the results in \textbf{Figure \ref{fig:change_fig1}}.

In addition to presenting the absolute change in solution, $|Q (\boldsymbol{\theta} - \boldsymbol{\theta}^{-i})|$, we also show the result of change in loss along each eigenvector, \textit{i.e.}, $L(\boldsymbol{\theta}^{-i})_j := \lambda_j (Q (\boldsymbol{\theta}^{-i} - \boldsymbol{\theta}))_j^2$ where $j$ is the index of the eigenvector. Note that this quantity is interesting because $L(\boldsymbol{\theta}^{-i}) = \sum_j L(\boldsymbol{\theta}^{-i})_j$. That is the loss for the sample solution can be decomposed into each individual eigenvector direction. Both plots show the  same trend of instability (largest change in solution and loss) along directions associated with small eigenvalues. Note that for the Fashion MNIST dataset, eigenvalue indices of $500$ and above have eigenvalues of $0$, making it unresponsive to swapping data because we use the minimum $L2$ norm solution for the degenerative case.

Given that fitting in directions associated with small eigenvalues could lead to poor generalization, we want to examine each methods' dependence on small eigenvalue directions. We quantify each optimizer's dependence on small eigenvalue directions by computing $\Vert P (\theta_{t+1} - \theta_{t}) \Vert_2 / \Vert (\theta_{t+1} - \theta_{t}) \Vert_2$, where $P$ is a projection onto eigenvectors associated with the largest (in absolute value) $10$ eigenvalues and $\theta_t$ represents the model parameters at iteration $t$ (a method proposed in \cite{ghorbani2019investigation}). We then average this value for all iterations. The higher the score, the less the model depends on directions associated with small eigenvalues. \textbf{Table \ref{table:dependence_on_large_eigenvalue_directions}} shows the results on ResNet18 for CIFAR-10. We observe that Adam depends on directions associated with small eigenvalues significantly more compared to AdaSGD and SGD (consistent with the result in \cite{gur2018gradient}).

\begin{table}[h]
  \caption{Methods' dependence on large eigenvalue directions for ResNet18 on CIFAR-10. Adam depends on directions associated with small eigenvalues significantly more compared to AdaSGD and SGD.}
  \centering
    \begin{tabular}{lllll}
    SGD & AdaSGD & Adam\\
    \hline
    0.0311 & 0.0502 & 0.0006
    \end{tabular} \label{table:dependence_on_large_eigenvalue_directions}
\end{table}




\section{Results on Additional Baselines, Datasets, and Architectures} \label{tiny_imagenet}

To further validate our findings, we included more baselines on MIMIC-III, CIFAR-10, and WikiText-2. In particular, we compare to \cite{li2018convergence} (a method that modifies the learning scheme of Adagrad instead of Adam), AdaSGDMax (provably convergent version of AdaSGD), and AMSgrad (provably convergent version of Adam). \textbf{Table \ref{table:result_convergent_baselines}} shows the results.

\begin{table*}[h]
  \caption{Comparison with additional baselines}
  \centering
    \begin{tabular}{lllll}
    Methods & MIMIC-III (AUC) & CIFAR-10 (accuracy) & WikiText-2 (perplexity)\\
    \hline
    AdaSGD & 0.849 & 94.00 & \textbf{109.96}\\
    \cite{li2018convergence} & 0.842 & 93.62 & 115.19\\
    AdaSGDMax & \textbf{0.851} & \textbf{94.06} & 110.96\\
    AMSgrad & 0.850 & 93.02 & 117.38\\
    \end{tabular} \label{table:result_convergent_baselines}
\end{table*}

Note that AdaSGD based methods are never performed worse than the baselines. In addition, we compared to baseline methods on the Tiny ImageNet dataset, ran a larger model on the CIFAR-100 dataset, and ran baselines on an unsupervised task on MNIST.

Following the results observed on CIFAR-10, WikiText-2, and MIMIC-3, we compare baseline methods on the Tiny ImageNet dataset, as ImageNet is a another common benchmark in computer vision. We used a ResNet18 architecture, adapting the ImageNet training code provided in the PyTorch examples repository\footnote{\url{https://github.com/pytorch/examples/tree/master/imagenet}}.

As the test set provided with Tiny ImageNet is unlabelled, we used Tiny ImageNet's validation set as our test set. We randomly fixed 20\% of the Tiny ImageNet training data as a validation set and used Top-1 accuracy on this validation set to search for the best $L2$ regularization strength and learning rate from the same ranges as our previous experiments. We randomly searched these ranges, training each pair of regularization strength and learning rate on one random seed. Based on when the validation accuracy and training loss plateaued during several initial test runs, we trained each setting for a maximum of 40 epochs and decayed the learning rate by a factor of 10 after 30 epochs. Given the best model for each optimizer, we reported the Top-5 accuracy on the test set in \textbf{Table \ref{table:tiny_imagenet_table}} and the top 1 result in \textbf{Table \ref{table:result_tinyimagenet_top1}}.

\begin{table*}[h]
  \caption{A comparison of optimization algorithms on Tiny ImageNet. Intervals listed are 95\% confidence intervals constructed by bootstrapping the test set.}
  \centering
    \begin{tabular}{cc}
    Methods & Tiny ImageNet Top-5 Test Accuracy \\
    \midrule
    SGD & 80.92 (80.28, 81.56) \\
    Adam & 80.16 (79.37, 80.95) \\
    AdaSGD & 81.34 (80.69, 81.99) \\
    Swats & 80.44 (79.66, 81.21) \\
    AdaBound & 80.59 (79.87, 81.30) \\
    \midrule
    \end{tabular} \label{table:tiny_imagenet_table}
\end{table*}

\begin{table}[h]
  \caption{Top 1 accuracies (\%) on TinyImagenet}
  \centering
    \begin{tabular}{lllll}
    SGD & AdaSGD & Adam & Swats & AdaBound\\
    \hline
    58.68 & 58.95 & 57.50 & 58.25 & 56.10
    \end{tabular} \label{table:result_tinyimagenet_top1}
\end{table}

To verify our findings on a larger model, we applied a 100 layer densenet to the CIFAR-100 dataset. Test accuracies (\%) are summarized in \textbf{Table \ref{table:result_deeper_network}}. AdaSGD outperforms transitional approaches and is competitive with SGD.

\begin{table}[h]
  \caption{Test accuracy (\%) for a 100-layer densenet on CIFAR-100. AdaSGD outperforms transitional approaches and is competitive with SGD.}
  \centering
    \begin{tabular}{lllll}
    SGD & AdaSGD & Adam & AdaBound & Swats\\
    \hline
    74.82 & 73.41 & 71.04 & 67.34 & 66.59
    \end{tabular} \label{table:result_deeper_network}
\end{table}

As an additional task, we applied a 2 layer variational autoencoder (VAE) on MNIST. \textbf{Table \ref{table:result_vae}} shows the test reconstruction error. Again AdaSGD outperforms transitional approaches.

\begin{table}[h]
  \caption{Test reconstruction error of a VAE on MNIST. AdaSGD outperforms transitional approaches.}
  \centering
    \begin{tabular}{lllll}
    SGD & AdaSGD & Adam & AdaBound & Swats\\
    \hline
    77.51 & 77.59 & 77.33 & 203.38 & 83.09
    \end{tabular} \label{table:result_vae}
\end{table}

\section{Theoretical properties} \label{properties}

Empirically, we observe that AdaSGD is robust to the choice of learning rate and has a similar optimization trajectory to SGD. In this section, we provide a theoretical analysis supporting those observations. For the same reason as Adam, AdaSGD does not converge with convex loss. Thus, to better understand properties of AdaSGD, we use the same trick introduced in \citealp{reddi2019convergence}, \textit{i.e.}, we analyze a close variant AdaSGDMax instead. Note - proofs remain unchanged with momentum, bias correction, and initialization, but those factors are excluded here for simplicity. We denote the stochastic gradient of the loss function at time $t$ as $\mathbf{g}_t$, model parameters at time $t$ as $\boldsymbol{\theta}_t$, the optimal solution as $\boldsymbol{\theta}^*$, and the number of parameters as $d$, AdaSGD updates as follows: $\boldsymbol{\theta}_{t+1} = \boldsymbol{\theta}_{t} - \eta_t \mathbf{g}_t$, $\eta_t = \eta / \sqrt{v_t / d}$, $v_t = \beta v_{t-1} + (1-\beta) \Vert \mathbf{g}_t \Vert^2_2$. In AdaSGDMax, we have $\eta_t = \eta / \sqrt{\hat{v}_t / d}$ and $\hat{v}_t = \max\{ \hat{v}_{t-1}, v_t \}$. In the case of constrained optimization, \textit{i.e.}, $\forall t$ $\boldsymbol{\theta}_t \in \mathcal{F} $ where $\mathcal{F}$ is some constrained set, we project $\boldsymbol{\theta}_t$ to $\Pi_{\mathcal{F}}(\boldsymbol{\theta}_t) := \min _{\boldsymbol{y} \in \mathcal{F}} \Vert \boldsymbol{\theta}_t - \boldsymbol{y}\Vert_2$ to satisfy the constraint. 

\textbf{Theorem 3.1} (\textbf{AdaSGDMax is robust to learning rate}).
\textit{
With strongly convex deterministic quadratic loss and bounded gradient, AdaSGDMax with $\beta_1 = 0$ converges for all $\eta>0$ except when $\lim_{t \to \infty}\eta_t = 2 / \lambda_{\max}$.
}

\begin{proof}
Denote error at time $t$ as $\boldsymbol{e}_t = \boldsymbol{\theta}_ t - \boldsymbol{\theta}^* $. We first derive the convergence range of SGD in the strongly convex deterministic (gradient descent instead of stochastic gradient descent) problem to be $(0, 2/\lambda_{\max})$. We then show that $\eta_t$ in AdaSGDMax converges to a value in $(0, 2/\lambda_{\max}]$. We then show that any value other than those on the boundary will lead AdaSGDMax to converge, since after reaching a value in $(0, 2/\lambda_{\max})$, AdaSGDMax behaves like SGD.

Consider an arbitrary strongly convex quadratic function $L(\boldsymbol{\theta}) = (\boldsymbol{\theta} - \boldsymbol{\theta}^*)^\intercal Q^\intercal \Lambda Q (\boldsymbol{\theta} - \boldsymbol{\theta}^*) + c$ where $Q$ is an orthonormal matrix, $\Lambda$ is diagonal matrix with maximum eigenvalue $\lambda_{\max}$ and minimum eigenvalue $\lambda_{\min} > 0$, and $c \in \mathbb{R}^d$ is an offset constant. Then we have 

\begin{align}
\boldsymbol{e}_{t+1} &= \boldsymbol{\theta}_ {t} - \eta_t \nabla L(\boldsymbol{\theta}_t) -  \boldsymbol{\theta}^*  - \eta_t \nabla L(\boldsymbol{\theta}^*)\\
&= \boldsymbol{\theta}_ {t} -  \boldsymbol{\theta}^* + \eta_t \nabla^2 L(\boldsymbol{z}) (\boldsymbol{\theta}_ {t} -  \boldsymbol{\theta}^* )
\end{align}

where $\boldsymbol{z}$ is some point in the feature space given by Taylor expansion. Therefore,

\begin{align}
    \boldsymbol{e}_{t+1} &= (I - \eta_t Q^\intercal \Lambda Q) \boldsymbol{e}_t
\end{align}

rearrange and get,

\begin{align}
    Q \boldsymbol{e}_{t+1} = (I-\eta_t \Lambda) Q \boldsymbol{e}_{t}
\end{align}

To converge, we just need $\eta_t > 0$ and $1-\eta_t \lambda_{\max} < -1$, that is $\eta_t \in (0, 2/\lambda_{\max})$. 

Note that $\eta_t$ is monotonically non-increasing because $\hat{v}_t$ cannot decrease due to the $\max$ operation. Furthermore, $\eta_t>0$ for all $t$. Thus, by the monotone convergence theorem, $\eta_t$ will converge. We denote the converging value $\eta^* := \lim_{t->\infty} \eta_t$.  Since the gradient by assumption is bounded, $\eta^* \neq 0$.  Then our task is to prove $\eta^* \in (0, 2/\lambda_{\max}]$.

Before going further, we set up notations to simplify the proof. Denote $\mathbf{\hat{e}} := Q \mathbf{\hat{e}}$, we have $\mathbf{\hat{e}}_{t+1, i} = (1-\eta_t \lambda_i)  \mathbf{\hat{e}}_{t,i}$ where $\lambda_i$   is the $i^{th}$ eigenvalue and $\mathbf{\hat{e}}_{t,i}$ is the error at time $t$ along the $i^{th}$ eigenvector directions where the indices follows the rule: $i<j \implies \lambda_i \geq \lambda_j$.  Without loss of generality, we assume $\mathbf{\hat{e}}_{1,1} \neq 0$,  which means that the error along the max eigenvalue direction is not vanishing. If this does not hold, we can reorder the indices, starting from the largest eigenvalue whose associated error is not 0. We further define $s = \{i: |1-\eta_t \lambda_i| \leq 1\}$as the set  of index with non expanding coefficients and $\overline{s} = \{i:|1-\eta_t \lambda_i| > 1\}$ as the set of index with expanding coefficients.

We first deal with the case where $\eta_t \in (0, 2/\lambda_{\max}]$. We show that in this case, for all $t'>t$, $\eta_{t'} = \eta_{t}$, \textit{i.e.}, $\eta^* \in (0, 2/\lambda_{\max}]$. Using the fact that in this case $|1-\eta_t \lambda_{\max}| \leq 1$, we have 

\begin{align}
\Vert \nabla L(\boldsymbol{\theta}_{t+1})  \Vert_2^2 &= \Vert \nabla L(\boldsymbol{\theta}_t - \eta_t \nabla L(\boldsymbol{\theta}_t)) \Vert_2^2\\
&= \Vert \nabla L (\boldsymbol{\theta}_t) -  \eta_t \nabla^2 L(\boldsymbol{z}) \nabla L(\boldsymbol{\theta}_t) \Vert_2^2\\
& = \Vert (I - \eta_t Q^\intercal \Lambda Q) \nabla L (\boldsymbol{\theta}_t) \Vert_2^2\\
&\leq \Vert (I - \eta_t \Lambda) \Vert_2^2 \Vert \nabla L (\boldsymbol{\theta}_t) \Vert_2^2\\
&< \Vert \nabla L (\boldsymbol{\theta}_t) \Vert_2^2
\end{align}
that is $v_t$ will decrease and thus $\hat{v}_t$ will stay the same, which means $\eta_t =\eta_{t+1}$. In other words, once AdaSGDMax enters SGD's rance of convergence, it will behave exactly like SGD and converge to the optimal solution thereafter. 

For the case where $\eta_t > 2 / \lambda_{\max}$, we prove by contradiction, assuming that $\eta^* > 2/\lambda_{\max}$, then $|1-\eta_t \lambda_{\max}| \geq |1-\eta^* \lambda_{\max}| > 1$. We have two cases:  $\Vert \mathbf{e}_{t+1} \Vert_2 > \Vert \mathbf{e}_{t} \Vert_2$ or  $\Vert \mathbf{e}_{t+1} \Vert_2  \leq \Vert \mathbf{e}_{t} \Vert_2$.  Note that $1 \in \overline{s}$ as otherwise we contradict the assumption.

\textbf{case 1:} $\Vert \mathbf{e}_{t+1} \Vert_2  \leq \Vert \mathbf{e}_{t} \Vert_2$ 

We show that we can only stay in this case for at most $\log_{|1-\eta^* \lambda_{\max}|} (\Vert \mathbf{e}_{t} \Vert_2 / |  \mathbf{\hat{e}}_{t,1} |)$ steps before transitioning to case 2. First, observe that staying in case 1 means error can at most be reduced by $\Vert  \mathbf{e}_t \Vert_2$. However, each time step will boost  error along the first eigenvector by at least $|1-\eta^* \lambda_{\max}|$. Note that error along the first eigenvector dimension is non zero because $\mathbf{\hat{e}}_{1,1} \neq 0$  and $1 \in \overline{s}$. Thus we want find $t'$ such that $\mathbf{\hat{e}}_{t+t',1} \geq |1-\eta^* \lambda_{\max}|^{t'} |\mathbf{\hat{e}}_{t, 1}|  \geq \Vert \mathbf{e}_t\Vert_2$. Solving the inequality gives the desired result.  
 
\textbf{case 2:} $\Vert \mathbf{e}_{t+1} \Vert_2 > \Vert \mathbf{e}_{t} \Vert_2$ 

We prove that we can only transition out of this case at most $t_1 = \ceil{\log_{1 + \frac{\eta_{t} (\lambda_1 -c)}  {\eta_{t} c - 1}} \{\frac{2 d \Vert \mathbf{\hat{e}}_t\Vert^2_{\infty}} {\mathbf{\hat{e}}_{t, 1}^2 \eta^* \lambda_1 (\eta^* \lambda_1 -2)}\}}$ times. The idea of the proof is to show that after $t_1$ steps, one additional update will increase $\mathbf{\hat{e}}_{t+t_1+1,i}$ from $\mathbf{\hat{e}}_{t+t_1,i}$ by an amount great enough that deduction in all other errors will not be enough to compensate.

\begin{align*}
\Vert \mathbf{e}_{t+t_1+ 1} \Vert_2^2 - \Vert \mathbf{e}_{t+t_1} \Vert_2^2 
&=\sum_{i=1}^d (\mathbf{\hat{e}}_{t+t_1 + 1, i}^2 - \mathbf{\hat{e}}_{t+t_1, i}^2) \\
&\geq \mathbf{\hat{e}}_{t+t_1 + 1, 1}^2 - \mathbf{\hat{e}}_{t+t_1, 1}^2 - \sum_{j\neq 1} \mathbf{\hat{e}}_{t+t_1, j}^2\\
&=((\eta_{t+t_1} \lambda_1 - 1)^2-1) \mathbf{\hat{e}}_{t+t_1, 1}^2  - \\
&\quad \mathbf{\hat{e}}_{t+t_1, 1}^2 - \sum_{j\neq 1} \mathbf{\hat{e}}_{t+t_1, j}^2\\
&=(\eta_{t+t_1} \lambda_1 (\eta_{t+t_1} \lambda_1 - 2)) \mathbf{\hat{e}}_{t+t_1, 1}^2 - \\ &\quad\sum_{j\neq 1} \mathbf{\hat{e}}_{t+t_1, j}^2
\end{align*}

The second step comes from the fact that error cannot be negative and thus the amount of decrease in error cannot be too large. We define $k = \max\{i : \lambda_i < \lambda_1\}$.  If $k$ does not exist, \textit{i.e.}, all eigenvalues are the same, then $t_1=0$ because every update will increase error along all dimensions and the error increment is at least $\Vert \mathbf{e}_{t+ t_1+\delta + 1} \Vert_2^2 - \Vert \mathbf{e}_{t+t_1 + \delta} \Vert_2^2 \geq (\eta^* \lambda_1 (\eta^* \lambda_1 - 2)) \Vert \mathbf{e}_{t+t_1 +\delta}^2 \Vert_2^2  \geq (\eta^* \lambda_1 (\eta^* \lambda_1 - 2)) (\eta^* \lambda_1 - 1)^{2\delta} \Vert \mathbf{e}_{t+t_1}^2 \Vert_2^2$. This also shows that the increment in error is exponential with time after $t+t_0$. If $k$ does exist, we define $c > 0$ such that  $\eta^* >  2 / c$ and $\lambda_k \leq c < \lambda_1$. $c$ must exist as $c  = \max \{  1/\eta^* + \lambda_1 / 2, \lambda_k\}$would satisfy the criteria. We introduce $c$ so that we can have a upper bound on the second term in the last equation. Expanding out the error along t, we have

\begin{align*}
\Vert \mathbf{e}_{t+t_1+ 1} \Vert_2^2 - \Vert \mathbf{e}_{t+t_1} \Vert_2^2 
&\geq (\eta_{t+t_1} \lambda_1 (\eta_{t+t_1} \lambda_1 - 2)) 
\cdot \\&\quad \Pi_{t'=0}^{t_1-1} (\eta_{t+t'} \lambda_1 - 1)^2 \mathbf{\hat{e}}_{t, 1}^2  \\&\quad - d \Vert \mathbf{\hat{e}}_t \Vert_\infty^2 \Pi_{t'=0}^{t_1-1} (\eta_{t+t'} c - 1)^2
\end{align*}

We want the righthand side to be greater than 0. However, for convenience later, we require the righthand to be greater than $\eta_{t+t_1}  \lambda_1 (\eta_{t+t_1} \lambda_1 -2) \mathbf{\hat{e}}^2_{t+t_1, 1} / 2$ where the $2$ in the denominator is arbitrary as long as it is greater than 1. That is, we want to find $t_1$ such that

\begin{align*}
\eta_{t+t_1}  \lambda_1 (\eta_{t+t_1} \lambda_1 -2) \mathbf{\hat{e}}^2_{t+t_1, 1} / 2 &< (\eta_{t+t_1} \lambda_1 (\eta_{t+t_1} \lambda_1 - 2)) \cdot \\&\quad \Pi_{t'=0}^{t_1-1} (\eta_{t+t'} \lambda_1 - 1)^2 \mathbf{\hat{e}}_{t, 1}^2 \\&\quad- d \Vert \mathbf{\hat{e}}_t \Vert_\infty^2 \Pi_{t'=0}^{t_1-1} (\eta_{t+t'} c - 1)^2  
\end{align*}

Rearrange the terms to get

\begin{align}
\Pi_{t'=0}^{t_1-1} (1 + \frac{\eta_{t+t'} (\lambda_1 -c)}  {\eta_{t+t'} c - 1})^2 > \frac{2 d \Vert \mathbf{\hat{e}}_t\Vert^2_{\infty}} {\mathbf{\hat{e}}_{t, 1}^2 \eta_{t+t_1} \lambda_1 (\eta_{t+t_1} \lambda_1 -2)} 
\end{align}
Note that each term in the lefthand side is a monotonically decreasing function with respect to $\eta_{t+t'}$ (prove this by showing that its derivative is negative) and that $\eta_t \geq \eta_{t+a}$  for all $a\geq 0$. We just need to find $t_1$ such that a lower bound of the lefthand side is greater than an upper bound of the righthand side. A lower bound of the lefthand side is

\begin{align}
\Pi_{t'=0}^{t_1-1} (1 + \frac{\eta_{t+t'} (\lambda_1 -c)}  {\eta_{t+t'} c - 1})^2 &\geq \Pi_{t'=0}^{t_1-1} (1 + \frac{\eta_{t} (\lambda_1 -c)}  {\eta_{t} c - 1})^2 \\
&=(1 + \frac{\eta_{t} (\lambda_1 -c)}  {\eta_{t} c - 1})^{2 t_1}  
\end{align}

An upper bound of the righthand side is
\begin{align}
\frac{2 d \Vert \mathbf{\hat{e}}_t\Vert^2_{\infty}} {\mathbf{\hat{e}}_{t, 1}^2 \eta_{t+t_1} \lambda_1 (\eta_{t+t_1} \lambda_1 -2)} \leq \frac{2 d \Vert \mathbf{\hat{e}}_t\Vert^2_{\infty}} {\mathbf{\hat{e}}_{t, 1}^2 \eta^* \lambda_1 (\eta^* \lambda_1 -2)} 
\end{align}
Combine the lower and upper bound, and take the log to get
\begin{align}
t_1 >  \log_{1 + \frac{\eta_{t} (\lambda_1 -c)}  {\eta_{t} c - 1}} \{\frac{2 d \Vert \mathbf{\hat{e}}_t\Vert^2_{\infty}} {\mathbf{\hat{e}}_{t, 1}^2  \eta^* \lambda_1 (\eta^* \lambda_1 -2)}\}
\end{align}
This means that after $t_1$ steps, we are guaranteed to get $\Vert \mathbf{e}_{t+ t_1+\delta + 1} \Vert_2^2 - \Vert \mathbf{e}_{t+t_1 + \delta} \Vert_2^2 > \eta_{t+t_1 + \delta}  \lambda_1 (\eta_{t+t_1 +\delta} \lambda_1 -2) \mathbf{\hat{e}}^2_{t+t_1+\delta, 1} / 2$  for $\delta \in \mathbb{N}$. Expanding along time we get

\begin{align*}
\Vert \mathbf{e}_{t+ t_1+\delta + 1} \Vert_2^2 - \Vert \mathbf{e}_{t+t_1 + \delta} \Vert_2^2 &> \eta_{t+t_1 + \delta}  \lambda_1 (\eta_{t+t_1 +\delta} \lambda_1 -2) \cdot \\&\quad \mathbf{\hat{e}}^2_{t+t_1+\delta, 1} / 2\\
&\geq \eta^*  \lambda_1 (\eta^* \lambda_1 -2) (\eta^* \lambda_1 -1)^{2\delta} \cdot \\&\quad \mathbf{\hat{e}}^2_{t+t_1, 1} / 2 \label{eq:grow_exp}
\end{align*}

The last equation shows that error is growing exponentially with respect to $\delta$.

Next, we show that in this case, $\eta_t$ will decrease by a respectable amount after a fixed iteration by first proving $\Vert \nabla L(\boldsymbol{\theta}_{t+1}) \Vert_2^2  > \Vert \nabla L(\boldsymbol{\theta}_t) \Vert_2^2$.

\begin{align*}
2 \Vert \nabla L(\boldsymbol{\theta}_{t+1}) \Vert_2^2  - 2 \Vert \nabla L(\boldsymbol{\theta}_t) \Vert_2^2 &= \Vert Q^\intercal \Lambda Q \mathbf{e}_{t+1} \Vert_2^2  \\&\quad - \Vert Q^\intercal \Lambda Q \mathbf{e}_{t} \Vert_2^2 \\
&=\mathbf{\hat{e}}_{t+1} \Lambda^2 \mathbf{\hat{e}}_{t+1} - \mathbf{\hat{e}}_{t} \Lambda^2 \mathbf{\hat{e}}_{t} \\
&= \sum_{i=1}^d (\mathbf{\hat{e}}^2_{t+1, i} - \mathbf{\hat{e}}^2_{t, i}) \lambda_i^2 \\
&= \sum_{i\in s} (\mathbf{\hat{e}}^2_{t+1, i} - \mathbf{\hat{e}}^2_{t, i}) \lambda_i^2 \\ &\quad+ \sum_{i \in \overline{s}} (\mathbf{\hat{e}}^2_{t+1, i} - \mathbf{\hat{e}}^2_{t, i}) \lambda_i^2\\
\end{align*}
Denote $\lambda_{imin} = \min \{ \lambda_i: i \in \overline{s} \}$ and $\lambda_{imax} = \max \{ \lambda_i: i \in s \}$. Observe that $\lambda_{imin} > \lambda_{imax}$ because only large eigenvalues can fall out of the convergence region. Thus using the fact that $\forall i \in s$, $\mathbf{\hat{e}}^2_{t+1, i} - \mathbf{\hat{e}}^2_{t, i}  \leq 0$ and  $\forall i \in \overline{s}$, $\mathbf{\hat{e}}^2_{t+1, i} - \mathbf{\hat{e}}^2_{t, i}  > 0$, we have 

\begin{align*}
2 (\Vert \nabla L(\boldsymbol{\theta}_{t+1}) \Vert_2^2  - \Vert \nabla L(\boldsymbol{\theta}_t) \Vert_2^2) &\geq \sum_{i\in s} (\mathbf{\hat{e}}^2_{t+1, i} - \mathbf{\hat{e}}^2_{t, i}) \lambda_{imax}^2 \\ &\quad + \sum_{i \in \overline{s}} (\mathbf{\hat{e}}^2_{t+1, i} - \mathbf{\hat{e}}^2_{t, i}) \lambda_{imin}^2\\
&\geq \sum_{i=1}^d (\mathbf{\hat{e}}^2_{t+1, i} - \mathbf{\hat{e}}^2_{t, i}) \lambda_{imin}^2\\
&=\lambda^2_{imin} (\Vert \mathbf{\hat{e}}_{t+1}\Vert_2^2  - \Vert \mathbf{\hat{e}}_t \Vert_2^2) \\
&=\lambda^2_{imin} (\Vert \mathbf{e}_{t+1}\Vert_2^2  - \Vert \mathbf{e}_t \Vert_2^2) \\
&> 0
\end{align*}

After $t_1$ steps, we are stuck in this case. Combine the last equation with the exponential growth of error difference, we have
\begin{align}
\Vert \nabla L(\boldsymbol{\theta}_{t + t_1 + \delta+1}) \Vert_2^2 - \Vert \nabla L(\boldsymbol{\theta}_{t + t_1 + \delta}) \Vert_2^2 \geq c_1 (\eta^*-1)^{2\delta}
\end{align}

where $c_1>0$ is a constant. Thus we have

\begin{align}
v_{t +t_1 +\delta + 1} &= \beta_2 v_{t+t_1 +\delta} + (1-\beta_2) \Vert \nabla L(\mathbf{\theta}_{t+t_1 +\delta})\Vert^2_2 \\
&\geq (1-\beta_2) c_1 (\eta^* -1)^{2(\delta-1)}\\
&= c_2 (\eta^* -1)^{2(\delta-1)}
\end{align}
Similarly, $\sqrt{\hat{v}_t}$ grows exponentially, which means that $\eta_t$ will decay exponentially eventually bringing $\eta_t \leq 2/ \lambda_{\max}$, contradicting the hypothesis.
\end{proof}

AdaSGDMax greatly expands the range of learning rate in which SGD converges. Note that in the converging range of SGD, AdaSGDMax share the same trajectory as SGD. The exceptional cases can be easily avoided by decreasing the learning rate once the loss stops going down. One should note that any learning rate decay scheme that have final learning rate bigger than 0 but way smaller than 1 for SGD will converge, regardless of the initial learning rate. A nice property about AdaSGDMax, as shown in the proof, is that it will not decrease learning rate once $\eta_t$ is in SGD's convergence range, preventing it from suffering slow convergence due to too small learning rate.



Next, we show that AdaSGDMax converges for the online optimization setting in \citep{reddi2019convergence,luo2019adaptive}. Sublinear convergence in this setting implies convergence in the empirical risk minimization setting. Please refer to \cite{reddi2019convergence} for a quick review of online optimization.

\textbf{Theorem 3.2} (\textbf{AdaSGDMax has sublinear regret}).
\textit{
Consider a constrained online optimization problem with a convex constraint set $\mathcal{F}$,  convex loss function $f_t$ with $\Vert \nabla f_t(\boldsymbol{\theta}) \Vert_\infty \leq G_\infty$ for all $t \in [T]$ and $\boldsymbol{\theta} \in \mathcal{F}$, we want to minimize regret $R_T := \sum_{i=1}^T f_t(\boldsymbol{\theta}_t) - \min_{\boldsymbol{\theta}^* \in \mathcal{F}} \sum_{i=1}^T f_t(\boldsymbol{\theta}^*)$. Assuming $\mathcal{F}$ has bounded diameter $D_\infty$ (\textit{i.e.,} $D_\infty = \max_{\boldsymbol{x}, \boldsymbol{y} \in \mathcal{F}} \Vert \boldsymbol{x} - \boldsymbol{y} \Vert_{\infty})$ and $\eta_t = \eta / \sqrt{t \hat{v}_t / d}$ (\textit{i.e.}, decay learning rate by $\sqrt{t}$), 
AdaSGDMax has regret 
$$R_T \leq \frac{D^2_\infty \sqrt{d \hat{v}_T T}}{2 \eta} + \frac{d^{3/2} G_\infty^2 \eta (2 \sqrt{T}-1)}{2 \sqrt{\hat{v}_1}}$$
}

\begin{proof}
By convexity, we have 
\begin{align}
R_T &= \sum_{i=1}^T f_t(\boldsymbol{\theta}_t) - \min_{\boldsymbol{\theta}^* \in \mathcal{F}} \sum_{i=1}^T f_t(\boldsymbol{\theta}^*)\\
&\leq \sum_{i=1}^T \mathbf{g}_t \cdot (\boldsymbol{\theta}_t - \boldsymbol{\theta}^*) \label{eq:two}
\end{align}

Using the update rule for $\boldsymbol{\theta}_{t+1}$, we have 

\begin{align}
\Vert \boldsymbol{\theta}_{t+1} - \boldsymbol{\theta}^* \Vert_2^2 &= \Vert \Pi_{\mathcal{F}} (\boldsymbol{\theta}_t - \eta_t \mathbf{g}_t) - \boldsymbol{\theta}^*  \Vert_2^2  \\
&\leq  \Vert \boldsymbol{\theta}_t - \eta_t \mathbf{g}_t  - \boldsymbol{\theta}^* \Vert_2^2 \quad \quad \quad \text{(convexity of $\mathcal{F}$)}\\ 
&= \Vert \boldsymbol{\theta}_t - \boldsymbol{\theta}^* \Vert_2^2 + \eta_t^2 \Vert \mathbf{g}_t \Vert_2^2 - 2 \eta_t \mathbf{g}_t \cdot (\boldsymbol{\theta}_t -  \boldsymbol{\theta}^*) 
\end{align}

Rearranging terms we have

\begin{align}
\mathbf{g}_t \cdot (\boldsymbol{\theta}_t -  \boldsymbol{\theta}^*) &\leq (\Vert \boldsymbol{\theta}_t - \boldsymbol{\theta}^*\Vert_2^2 - \Vert \boldsymbol{\theta}_{t+1} - \boldsymbol{\theta}^* \Vert_2^2) / (2 \eta_t) \nonumber \\ &\quad + \eta_t \Vert \mathbf{g}_t \Vert_2^2 / 2\\
&\leq (\Vert \boldsymbol{\theta}_t - \boldsymbol{\theta}^*\Vert_2^2 - \Vert \boldsymbol{\theta}_{t+1} - \boldsymbol{\theta}^* \Vert_2^2) / (2 \eta_t) \nonumber \\ &\quad + \eta_t d G_\infty^2 / 2 \label{eq:sparse1}
\end{align}

Combined with Equation \ref{eq:two}, we have

\begin{align*}
R_T &\leq \sum_{i=1}^T \mathbf{g}_t \cdot (\boldsymbol{\theta}_t - \boldsymbol{\theta}^*)\\
&\leq \sum_{i=1}^T (\Vert \boldsymbol{\theta}_t - \boldsymbol{\theta}^*\Vert_2^2 - \Vert \boldsymbol{\theta}_{t+1} - \boldsymbol{\theta}^* \Vert_2^2) / (2 \eta_t) \nonumber \\ &\quad + \frac{d G^2_\infty}{2} \sum_{i=1}^T \eta_t \\
&\leq \Vert \boldsymbol{\theta}_1 - \boldsymbol{\theta}^* \Vert_2^2 / (2 \eta_1) + \sum_{i=2}^T \Vert \boldsymbol{\theta}_t - \boldsymbol{\theta}_* \Vert_2^2 (1 / (2\eta_t) \nonumber \\ &\quad - 1 / (2 \eta_{t-1})) + \frac{d G^2_\infty}{2} \sum_{i=1}^T \eta_t\\
\end{align*}
Using the fact that $\eta_t$ monotonically decreases, we have

\begin{align}
R_T &\leq \frac{d D_\infty^2}{2\eta_1} + \frac{d D_\infty^2}{2} \sum_{i=2}^T(1/\eta_t - 1/\eta_{t-1}) + \frac{d G^2_\infty}{2} \sum_{i=1}^T \eta_t \label{eq:sparse2} \\
&= \frac{d D_\infty^2}{2 \eta_T} + \frac{d G^2_\infty}{2} \sum_{i=1}^T \eta_t
\end{align}

Using the definition of $\eta_t$
\begin{align}
R_T &\leq \frac{D^2_\infty \sqrt{d \hat{v}_T T} }{2 \eta} + \frac{d G_\infty^2}{2} \sum_{i=1}^T \eta / \sqrt{t \hat{v}_t / d}\\
&= \frac{D^2_\infty \sqrt{d \hat{v}_T T}}{2 \eta} + \frac{d G_\infty^2 \eta \sqrt{d}}{2} \sum_{i=1}^T  1 / \sqrt{t \hat{v}_t} \\
&\leq \frac{D^2_\infty \sqrt{d \hat{v}_T T}}{2 \eta} + \frac{d G_\infty^2 \eta \sqrt{d}}{2} \sum_{i=1}^T  1 / \sqrt{t \hat{v}_1} \\
&\leq \frac{D^2_\infty \sqrt{d \hat{v}_T T}}{2 \eta} + \frac{d^{3/2} G_\infty^2 \eta (2 \sqrt{T}-1)}{2 \sqrt{\hat{v}_1}}
\end{align}
\end{proof}

\textbf{Corollary 3.2.1}
\textit{
Setting $\eta_t = \eta D_\infty / (G_\infty \sqrt{t \hat{v}_t })$, we have
$$R_T \leq \frac{d D_\infty G_\infty \sqrt{\hat{v}_T T}}{2 \eta} + \frac{d D_\infty G_\infty \eta (2 \sqrt{T}-1)}{2 \sqrt{\hat{v}_1}}$$
}

This bound is comparable to SGD's regret bound of $\frac{d D_\infty G_\infty \sqrt{T}}{2\eta} + \frac{d D_\infty G_\infty \eta(2\sqrt{T}-1)}{2}$.


\textbf{Theorem 3.3} (\textbf{AdaSGD reaches a solution close to the optimal solution}).
The distance from AdaSGD's solution $\boldsymbol{\theta}$ to the optimal solution $\boldsymbol{\theta}^*$ for deterministic strongly convex quadratic problems with $\eta >0$ and bounded gradient is bounded by $\Vert \boldsymbol{\theta} - \boldsymbol{\theta}^* \Vert_2 \leq \frac{\sqrt{d} \eta \mathcal{K}} { 2 (1-\beta_2)}$, where $\mathcal{K}=\lambda_{\max} / \lambda_{\min}$ is the condition number. \label{theorem:close_to_sln}

\begin{proof}
Consider an arbitrary strongly convex quadratic function $L(\boldsymbol{\theta}) = 0.5 (\boldsymbol{\theta} - \boldsymbol{\theta}^*)^\intercal Q^\intercal \Lambda Q (\boldsymbol{\theta} - \boldsymbol{\theta}^*) + c$ where $Q$ is an orthonormal matrix, $\Lambda$ is diagonal matrix with maximum eigenvalue $\lambda_{\max}$ and minimum eigenvalue $\lambda_{\min} > 0$, and $c \in \mathbb{R}^d$ is an offset constant. We have $\nabla L(\boldsymbol{\theta}_t) = Q^\intercal \Lambda Q (\boldsymbol{\theta}_t -\boldsymbol{\theta}^*)$. Rewriting it, we have
\begin{align}
\Vert \boldsymbol{\theta}_t - \boldsymbol{\theta}^* \Vert_2  &\leq \Vert \Lambda^{-1} \Vert_2 \Vert  \nabla L(\boldsymbol{\theta}_t)\Vert_2\\
&= 1/\lambda_{\min} \Vert  \nabla L(\boldsymbol{\theta}_t)\Vert_2 \label{eq:error}
\end{align}
We then bound $\Vert  \nabla L(\boldsymbol{\theta}_t)\Vert_2$ by considering two cases (note that $\eta_t \neq 0$ because $\eta \neq 0$ and the gradient is bounded). In case 1, $\eta_t < 2/\lambda_{\max}$, the error will keep decreasing because it is in the converging range of SGD.   In case 2, $\eta_t \geq 2/\lambda_{\max}$. Expanding out the definition of $\eta_t$ and rearrange, we get $v_t \leq (\eta \lambda_{\max} /2) ^2 d$. Since $v_t = \beta_2 v_{t-1} + (1-\beta_2) \Vert  \nabla L(\boldsymbol{\theta}_t)\Vert_2^2$, we have  $\Vert  \nabla L(\boldsymbol{\theta}_t)\Vert_2 < \frac{\sqrt{d} \eta \lambda_{\max}}{2(1-\beta_2)}$. Thus combined with  Equation \ref{eq:error}, we have

\begin{align}
\Vert \boldsymbol{\theta} - \boldsymbol{\theta}^* \Vert_2 \leq \frac{\sqrt{d} \eta \mathcal{K}} { 2 (1-\beta_2)}
\end{align}
\end{proof}

Empirically, AdaSGD performs better than AdaSGDMax. The only difference between AdaSGD and AdaSGDMax is that in the former the learning rate is allowed to decrease. Empirically, this leads to good solutions faster, since the algorithm can accelerate in flat regions and slow down when the gradient changes quickly. 

\end{document}